\documentclass{article}

\usepackage{microtype}
\usepackage{graphicx}
\usepackage{subcaption}
\usepackage{booktabs}
\usepackage{hyperref}
\usepackage{url}

\usepackage{amsmath}
\usepackage{amssymb}
\usepackage{mathtools}
\usepackage{enumitem}
\usepackage{pifont}
\usepackage{threeparttable}
\usepackage{wrapfig}
\usepackage{verbatim}

\usepackage[preprint]{icml2026}
\makeatletter
\renewcommand{\ICML@preprint}{\textit{Accepted as an oral presentation at the GFM @ ICML 2026 Workshop.}}
\makeatother

\usepackage[capitalize,noabbrev]{cleveref}

\newcommand{\State}{\STATE}
\newcommand{\Require}{\REQUIRE}
\newcommand{\Ensure}{\ENSURE}
\newcommand{\For}{\FOR}
\newcommand{\EndFor}{\ENDFOR}
\newcommand{\If}{\IF}
\newcommand{\Else}{\ELSE}
\newcommand{\EndIf}{\ENDIF}
\newcommand{\Return}{\STATE \textbf{return} }

\icmltitlerunning{GILT: LLM-Free, Tuning-Free Graph In-Context Learning}

\begin{document}

\twocolumn[
  \icmltitle{GILT: An LLM-Free, Tuning-Free Graph Foundational Model for In-Context Learning}

  % TODO: Replace the placeholder author block below with the final public author list.
  \begin{icmlauthorlist}
    \icmlauthor{Weishuo Ma}{wict}
    \icmlauthor{Yanbo Wang}{ai}
    \icmlauthor{Xiyuan Wang}{ai}
    \icmlauthor{Lei Zou}{wict}
    \icmlauthor{Muhan Zhang}{ai}
  \end{icmlauthorlist}

  \icmlaffiliation{ai}{Institute for Artificial Intelligence, Peking University}
  \icmlaffiliation{wict}{Wangxuan Institute of Computer Technology, Peking University}

  \icmlcorrespondingauthor{Muhan Zhang}{muhan@pku.edu.cn}

  \icmlkeywords{Graph Foundation Models, In-Context Learning, Few-Shot Learning, Graph Neural Networks}

  \vskip 0.3in
]

\printAffiliationsAndNotice{}
\begin{abstract}
Graph Neural Networks (GNNs) are powerful tools for processing relational data but often struggle to generalize to unseen graphs, giving rise to the development of Graph Foundational Models (GFMs). However, current GFMs are challenged by the extreme heterogeneity of graph data, where each graph can possess a unique feature space, label set, and topology. To address this, two main paradigms have emerged. The first leverages Large Language Models (LLMs), but is fundamentally text-dependent, thus struggles to handle the numerical features in vast graphs. The second pre-trains a structure-based model, but the adaptation to new tasks typically requires a costly, per-graph tuning stage, creating a critical efficiency bottleneck. In this work, we move beyond these limitations and introduce \textbf{G}raph \textbf{I}n-context \textbf{L}earning \textbf{T}ransformer (GILT), a framework built on an LLM-free and tuning-free architecture. GILT introduces a novel token-based framework for in-context learning (ICL) on graphs, reframing classification tasks spanning node, edge and graph levels in a unified framework. This mechanism is the key to handling heterogeneity, as it is designed to operate on generic numerical features. Further, its ability to understand class semantics dynamically from the context enables tuning-free adaptation. Comprehensive experiments show that GILT achieves stronger few-shot performance with significantly less time than LLM-based or tuning-based baselines, validating the effectiveness of our approach. Our code is available at: \url{https://github.com/yiming421/inductnode/}.
\end{abstract}

\section{Introduction}

Graph Neural Networks (GNNs) have emerged as the standard for processing graph data, achieving state-of-the-art performance on a wide range of single-graph tasks \citep{gcn,vgae,gat,gin}. However, their fundamental limitation is a lack of generalization: a GNN trained on one graph often fails to transfer to an unseen graph with different features or topology \citep{hu2020}. In parallel, the artificial intelligence area has been reshaped by the success of foundation models that exhibit remarkable transferability in domains like language \citep{gpt3} and vision \citep{clip}. This confluence of GNNs' limitations and the power of the foundation model paradigm has spurred intense interest in a new frontier: the Graph Foundational Model (GFM).

However, the extreme heterogeneity in graph data presents a fundamental obstacle to realizing this vision. Unlike text or images which benefit from a universal vocabulary, graphs lack this common foundation. Each graph can possess its own arbitrary feature space, with varying dimensions and semantics; a distinct target space with its own unique structure like discrete classes or continuous values; and a vastly different topological structure \citep{mao24}. Consequently, the parameters of a conventional GNN architecture are fundamentally tied to the specific feature and output formats of its training data, making the model inherently non-transferable \citep{hu2020}. This core challenge of bridging graph heterogeneity has driven the development of the main GFM paradigms to date \citep{liu25}.

% The first primary paradigm in constructing GFMs leverages Large Language Models (LLMs) to solve the heterogeneity problem by unifying all heterogeneous data within the language space \citep{li24,fan24,ren24}. The core strategy involves applying an LLM to interpret the textual information associated with a graph's nodes and edges \citep{ren24}. Although highly effective for text-rich graphs like citation networks, this method's applicability is fundamentally constrained to text-attributed data \citep{graphany,gcope,riemanngfm,samgpt}. Consequently, the paradigm is unsuitable for graphs with numerical, categorical, or purely structural data, as is common in fields like molecular biology \citep{moleculenet}.

The first primary paradigm tackles the heterogeneity problem by leveraging Large Language Models (LLMs) to create a unified semantic space \citep{li24,fan24,ren24}. The core strategy involves applying an LLM to interpret the textual information associated with a graph's nodes and edges, mapping diverse features and labels into a shared space. Although effective for text-rich graphs like citation networks, this approach introduces a dependency on textual data \citep{graphany,gcope,riemanngfm}. Consequently, it is unsuitable for graphs with numerical, categorical, or purely structural data, as is common in fields like molecular biology \citep{moleculenet}.

% The second major paradigm known as graph prompting takes a more direct approach to heterogeneity by real-time parameter adaptation, typically involving pre-training a GNN encoder on large-scale data and then adapting it to downstream tasks \citep{all_in_one,riemanngfm,samgpt}. While this approach avoids the text-dependency issue, its primary mode of adaptation introduces a significant tuning barrier. To solve a new task, these models almost always require a round of gradient-based updates. Although methods have evolved from costly full fine-tuning to more parameter-efficient strategies like prompt-tuning, the fundamental need to modify model weights for each specific graph or task persists \citep{sun23}. This incurs considerable computational costs and diverges from the promise of a foundational model: a ready-to-use system that can be applied out-of-the-box \citep{graphany}.

The second major paradigm, known as graph prompting, takes a more direct and graph-native solution to heterogeneity by real-time parameter adaptation, typically involving pre-training a GNN encoder on large-scale data and then adapting it to downstream tasks \citep{all_in_one,riemanngfm,samgpt}. While this approach avoids the text-dependency issue, it introduces a dependency on tuning. The need to modify model weights for each new graph or task persists \citep{sun23}, creating a significant efficiency bottleneck and diverging from the promise of a truly ``out-of-the-box'' foundational model \citep{graphany}. 

% To overcome both the text and tuning barriers simultaneously, we introduce the \textbf{Graph In-context Learning Transformer (GILT)}, a framework centered on the powerful paradigm of In-Context Learning (ICL) to enable both \textbf{LLM-free and tuning-free}. GILT's architecture first uses a graph-native encoder to convert an arbitrary few-shot task into a standardized set of tokens. These tokens are then processed by a specialized ICL Transformer for contextual reasoning, and a final prediction is made by a tuning-free Prototypical Head. This design is key to overcoming the core challenge of an LLM-free and tuning-free GFM. By removing the reliance on the shared natural language, the feature and target spaces become arbitrary and heterogeneous across graphs, a problem traditionally solved with test-time parameter update. GILT's in-context learning ability is the crucial element that bypasses this need. It allows the model to dynamically understand the semantics of any new feature and label space during inference, using only the context provided by the prompted examples.

In this work, we move beyond the two barriers by introducing the \textbf{Graph In-context Learning Transformer (GILT)}, a framework designed to be both \textbf{LLM-free and tuning-free}. Our key innovation is to reframe few-shot graph tasks, spanning node, edge, and graph classification, as a unified token-based in-context learning problem. GILT's architecture first tokenizes a task, converting its structure and features into a set of tokens. These tokens are then processed by a Transformer that learns the task's semantics directly from the prompted examples. This in-context learning mechanism allows GILT to dynamically interpret unseen feature and label spaces at inference time, completely bypassing the need for textual information or parameter updates.

% We conducted extensive experiments to validate our approach across a diverse suite of unseen graphs from multiple domains, spanning node classification, link prediction, and graph classification tasks. The results demonstrate that a pre-trained GILT model achieves strong few-shot performance. Our framework not only shows competitive performance against state-of-the-art GFMs but also proves to be a more general and efficient solution. Specifically, It demonstrates strong performance on graphs with pure numeric feature where LLM-dependent baselines are inapplicable, and delivers its predictions significantly faster than methods requiring per-graph tuning. Furthermore, comprehensive ablation studies validate the importance of our core architectural choices. These findings validate that the in-context learning paradigm is a powerful path toward universal GFMs.

% The experiments summary should be aligned with experiment part.

% We validate GILT through comprehensive experiments on multiple benchmarks, demonstrating that it achieves strong few-shot performance on a diverse suite of unseen graphs. Our framework not only proves competitive with state-of-the-art GFMs, but also offers a more general and efficient solution, delivering strong performance on numerical graphs where LLM-based methods are inapplicable and generating predictions significantly faster than tuning-based approaches. % We also conducted ablation studies on xxx, xxx, revealing the potential of in-context learning paradigm to a universal GFM. (Here may depending on later experiments)

We validate GILT on diverse benchmarks spanning node, link, and graph classification, with results confirming its state-of-the-art few-shot performance. Its LLM-free design allows it to operate directly on text-independent graphs with numerical features, where text-based models are often either inapplicable or require laborious pre-processing to manually create textual descriptions. Concurrently, its tuning-free nature provides a significant efficiency advantage, making it orders of magnitude faster than methods that require per-graph tuning and positioning it as a more scalable solution. Our code is available at: \url{https://github.com/yiming421/inductnode/}.

Our main contributions in this work are as follows:
\vspace{-1em}

\begin{itemize}[leftmargin=10pt]
% \item We propose a novel paradigm for Graph Foundational Models centered on LLM-free, tuning-free In-Context Learning, reframing few-shot graph problems as a universal token-reasoning task.
\item We design and implement GILT, a LLM-free, tuning-free In-Context Learning architecture, reframing few-shot graph problems as a unified token-reasoning task.

\vspace{-0.6em}

\item We collect diverse datasets for pretraining and trained one model for multiple tasks through a graph-native tokenization pipeline and a two-stage ICL Transformer.

\vspace{-0.6em}
\item We provide comprehensive empirical validation establishing GILT's superiority. Our experiments demonstrate its state-of-the-art few-shot performance in text-free graphs, and its efficiency on faster than both tuning-based adaptation and the heavy inference required by LLMs. 
% \item We propose a new paradigm for Graph Foundational Models that is both LLM-free and tuning-free which reframes few-shot learning on graphs as a pure in-context learning task.
% Possibly we can have a 3-rd contribution on comprehensive ablations, showing our selection on PCA, linear GCN, pre-training are positive for generalization. 
% \item We provide extensive empirical evidence that a single, pre-trained GILT model achieves strong and competitive few-shot performance across a diverse suite of benchmarks for node, link, and graph classification, validating our approach as a more general and efficient path toward a universal GFM.

% \item We design and implement GILT, a novel architecture centering LLM-free, tuning-free In-Context Learning, featuring a graph-native tokenization pipeline and a specialized two-stage ICL Transformer with a Prototypical Head.
% \item Through experiments on diverse benchmarks, we show that GILT achieves strong few-shot performance across unseen graph, and we conduct comprehensive ablation studies to validate our principled design choices.

\end{itemize}

\section{Related Work}

The development of GFMs can be understood through two lenses: \textbf{the core techniques} used in its architecture and \textbf{the target learning paradigm}, which defines how a model adapts to new tasks. This section reviews the field along these lines, showing how limitations of current techniques motivate a paradigm shift towards in-context learning.

\begin{table*}[!t]
% \vspace{-0.5em}
\caption{Comparison of GILT with representative Graph Foundational Model paradigms.}
\label{tab:comparison}
\vspace{-0.5em}
\centering
\begin{threeparttable}
\resizebox{0.92\textwidth}{!}{%
\begin{tabular}{@{}lcccccc@{}}
\toprule
\textbf{Method / Paradigm} & \textbf{LLM-Free} & \textbf{Tuning-Free}  & \begin{tabular}[c]{@{}c@{}}\textbf{Multi-Domain}\\\textbf{Pre-training}\end{tabular} & \textbf{Multi-Task} & \textbf{Few-shot} \\
\midrule
ZeroG & \ding{55} & \ding{51} & \ding{51} & \ding{55} & \ding{55}  \\
GOFA & \ding{55} & \ding{51} & \ding{51} & \ding{51} & \ding{55}  \\
GCOPE & \ding{51} & \ding{55} & \ding{51} & \ding{51} & \ding{51} \\
RiemannGFM & \ding{51} & \ding{55} &  \ding{51} & \ding{51} & \ding{51} \\
OFA & \ding{55} & \ding{51} & \ding{51} & \ding{51} & \ding{51} \\
GraphAny & \ding{51} & \ding{51} & \ding{55} & \ding{55} & \ding{51} \\
\midrule
\textbf{GILT (Ours)} & \textbf{\ding{51}} & \textbf{\ding{51}} & \textbf{\ding{51}} & \textbf{\ding{51}} & \textbf{\ding{51}} \\
\bottomrule
\end{tabular}%
}
\end{threeparttable}
\vspace{-1.5em}
\end{table*}

\subsection{Popular Techniques in GFM Architecture}

To address key challenges like graph heterogeneity and task adaptation, researchers have developed several powerful techniques that are often used in combination.

% The pursuit of a GFM has led to the development of two primary research tracks, each with distinct approaches to handling the heterogeneity of graph data. In this section, we review these paradigms and then situate our work within the emerging field of in-context learning.

\textbf{LLMs for GFMs}
A primary research direction for building GFMs involves unifying heterogeneous graph data within the text domain to leverage the capabilities of LLMs. These efforts can be broadly grouped into two strategies. The first uses an \textbf{LLM as an enhancer}, processing diverse textual node features into a common representation space before performing structure-aware prediction \citep{prodigy,ofa,chen23,zerog,tape,gft,plenz24,graphcli,opengraph,llmbp}. For instance, ZeroG \citep{zerog} leverages an LLM to encode textual node attributes into a unified semantic space and then performs neighborhood aggregation for the final prediction. The second strategy employs the \textbf{LLM as the predictor}, aiming to leverage its generative capabilities for greater task flexibility \citep{graphgpt,llaga,unigraph,graphtranslator,gofa,askgnn,unigraph2,teagfm,instructg,graphtranslate,graphicl}. The GOFA model \citep{gofa} achieves this by interleaving GNN layers into an LLM's architecture, combining message-passing with semantic reasoning before generating prediction. Ultimately, the reliance of both strategies on a textual foundation naturally limits their scope to text-attributed graphs.

\textbf{Graph Prompting}
% A parallel line of work focuses on creating graph-native models that do not depend on LLM. 
Another dominant technique is graph prompting, which introduces learnable structural components that steer a pre-trained model's reasoning across diverse tasks. Following an initial large-scale pre-training phase, the GNN's weights are frozen. Adaptation is then achieved by introducing and tuning small, learnable prompts that adapt the model's behavior for new tasks \citep{gppt,graphprompt,all_in_one,gft2,prog,gcope,riemanngfm,fug,samgpt,relief,mdpgnn,csgnn,dagprompt,mdgfm,graphlora,bridge}. For instance, the
GCOPE \citep{gcope} framework extends a prompt-like mechanism to the pre-training stage, using learnable coordinators as virtual nodes to align different graph datasets. RiemannGFM \citep{riemanngfm} introduces a novel geometric perspective, pre-training a model on a universal ``structural vocabulary'' of trees and cycles, which is then adapted to new tasks through prompt tuning. While parameter-efficient, these methods' adaptation process still hinges on \textbf{gradient-based updates for each new graph}.

% The calling of "graph prompting".

\subsection{The Paradigm Shift Towards In-Context Learning}

The ultimate goal for a GFM is to operate as a ready-to-use system that generalizes to new tasks without re-training. This has motivated a shift towards In-Context Learning, a paradigm where a pre-trained model solves a new task at inference time using only a few prompted examples, without any parameter updates \citep{gpt3}. While popularized by LLMs, the success of ICL on structured tabular data \citep{tabpfnv1,tabpfn}, computer vision \citep{iclcv}, and time-series \citep{icltime} has underscored its potential for more modalities. 

Pioneering frameworks have begun to explore this direction. OFA \citep{ofa}, for instance, enables in-context learning through constructing a unified prompt graph connecting labeled support examples to their class nodes, allowing a GNN to synthesize this structural context for classification in a single forward pass. More recently, GraphAny \citep{graphany} achieved tuning-free generalization by using a pre-trained attention module to fuse the outputs of multiple non-parametric, analytical solvers. These works established the viability of tuning-free adaptation on graphs and laid the groundwork for more advanced ICL systems.

Building on these insights, GILT introduces a more general and powerful framework for in-context learning on graphs which uses a Transformer backbone chosen for its proven strength in ICL across language and tabular domains. With the aid of a specialized graph encoding module, the framework is inherently LLM-free to learn from raw numerical features alone.  Its deep, pre-trained model is designed to learn complex non-linear patterns, making it a flexible solution for node, link, and graph classification tasks.

% \small % Start a small-font block for the notes
% \noindent % Prevent indentation
% \textsuperscript{1} While the principles are general,  GCOPE is only evaluated on node classification. \\
% \textsuperscript{2} Although its analytical mechanism is inherently capable of few-shot inference, GraphAny is presented as a fully-inductive model for semi-supervised tasks and lacks evaluation under the few-shot setting.
\section{Method}

To overcome the text and tuning barriers inherent in current GFMs, we introduce the GILT framework. The core technical contribution of our work is to reframe the few-shot graph learning task as a problem of reasoning over a set of contextual tokens, allowing us to leverage the power of the Transformer for universal in-context reasoning on graphs.

We formally define a graph as $G = (V, E)$ with node features $X \in \mathbb{R}^{|V| \times d_{in}}$ and an adjacency matrix $A$. GILT is designed for the \textbf{few-shot, in-context learning} setting. Each task is an $N$-way $K$-shot problem, where the model is given a \textbf{support set} of labeled examples, $\mathcal{S} = \{(x_i, y_i)\}_{i=1}^{N \times K}$, and must predict labels for an unseen \textbf{query set}, $\mathcal{Q} = \{x_j\}_{j=1}^{Q}$. The items $x_i$ can be nodes, edges, or entire graphs. The model must leverage the context from $\mathcal{S}$ to make predictions for $\mathcal{Q}$ without any parameter updates.

\subsection{Architecture Overview}

\begin{figure*}[t]
    \centering
    \includegraphics[width=0.88\textwidth]{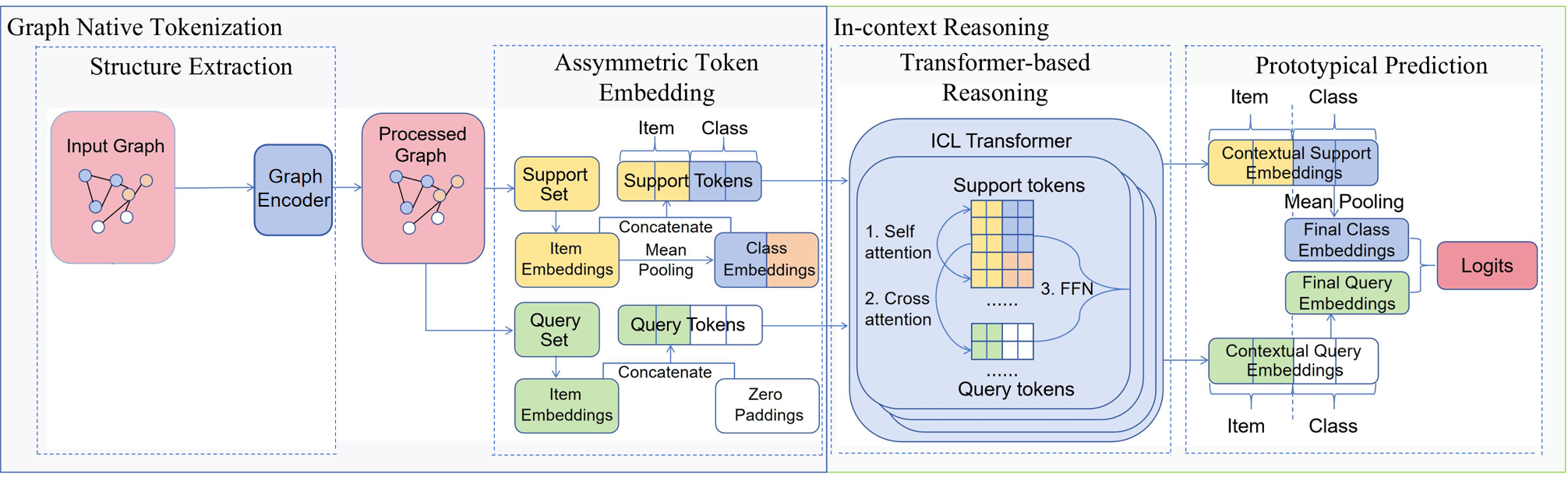}
    \caption{GILT begins with a graph-native tokenization module converting a few-shot task into unified tokens. This module first uses a GNN to generate structure-aware embeddings, which are then combined with class prototypes to form the support and query tokens. The tokens are then passed to ICL Transformer, which features a two-stage attention mechanism for in-context reasoning and a Prototypical Head for the final classification.}
    \label{fig:model}
    \vspace{-1.5em}
\end{figure*}

As illustrated in Figure~\ref{fig:model}, GILT is a two-phase pipeline designed to first translate graph tasks into a universal format and then reason over that format to make predictions.

\textbf{Phase 1: Graph-Native Tokenization (Syntactic Unification).} The first phase tackles graph heterogeneity. It converts a raw, few-shot graph task with its unique features and structure into a standardized set of contextual tokens. This creates a unified format that represents the problem.

\textbf{Phase 2: In-Context Reasoning (Semantic Unification).} The second phase is designed to understand the task's meaning from the tokens alone without any tuning. A specialized ICL Transformer processes the token set, learning the task's rules from the provided examples to make a prediction.

\subsection{Graph-Native Tokenization}

The Graph-Native Tokenization phase performs \textbf{syntactic unification}, converts raw graphs and the few-shot task definition into the set-based representation required by the ICL reasoning phase. This section details the components of this phase: the extraction of structural information, and the final task-specific asymmetric token formulation.

% \textbf{Initial Graph Encoding.}
\textbf{Extraction of Structure Information.}
The tokenization pipeline begins with a structural encoder to inject topological context before forming task-specific tokens.
The processed raw feature matrix is enriched with topological information by our structural encoder, a deep, multi-layer Graph Convolutional Network (GCN). Critically, we employ a \textbf{linear version of the GCN}, omitting the learnable weight matrices and non-linear activations. This design aligns with simplified GCN architectures such as SGC \citep{sgc} and APPNP \citep{appnp}, which demonstrate that stripping away non-linearities can mitigate overfitting while effectively capturing structural information.
The rationale for this choice is rooted in our design principle: a learnable projection at this stage tends to overfit the feature semantics of the pre-training graphs, hindering generalization. By using a simple, parameter-free aggregator, we ensure its role is to strictly extract local structural patterns, deferring all complex semantic reasoning to the more powerful ICL Transformer. 
We also find that a deeper encoder (4-6 layers) provides a richer, multi-hop context that is highly beneficial for the downstream Transformer.
% The #layer is also a very strong assumption, should with ablation
% While traditional supervised GCNs often peak at 2-3 layers, \textbf{we find that a deeper 4-6 layer linear encoder is significantly more effective in our framework}. We attribute this to the downstream Transformer benefiting from the richer, multi-hop structural context gathered by a deeper aggregator. 
To stabilize the representations as they propagate through the deep encoder, each linear aggregation is followed by an independent LayerNorm with its own learnable affine parameters:

\vspace{-0.5em}

\begin{equation}
    H^{(l+1)} = \text{LayerNorm}(\tilde{A}H^{(l)})
\end{equation}

\vspace{-0.5em}

where $\tilde{A}$ is the normalized adjacency matrix with self-loops and $H^{(0)} = X'$. The output of the final layer is the node embedding matrix $H \in \mathbb{R}^{|V| \times d}$.

\textbf{Asymmetric Token Formulation.} 
The next step is to encode the entire N-way K-shot task into a set of fixed-dimension tokens. We derive a single vector representation, denoted $h$, for each task item (a node's embedding for node tasks, an element-wise product for link tasks, or a pooled vector for graph tasks).
% Following the initial graph encoding, we construct a single vector representation for each task item based on its granularity: the representation is the node's own embedding for a \textbf{node} item; the element-wise product of its endpoint embeddings for a \textbf{link} item; and a pooled vector for a \textbf{graph} item.
These item representations, $h_i$, are then used to form the final tokens through an asymmetric process. An initial representation for each class $p_c$ is computed using simple \textbf{mean pooling} over the support item representations, followed by L2 normalization for stability. This class representation is then paired with each support item's representation to form the final support tokens, while query item representations are paired with zero-padding:
\begin{equation}
\begin{aligned}
\mathbf{p}_c &= \frac{\frac{1}{|\mathcal{S}_c|}\sum_{(x_i, y_i) \in \mathcal{S}_c} \mathbf{h}_i}{\left\| \frac{1}{|\mathcal{S}_c|} \sum_{(x_i, y_i) \in \mathcal{S}_c} \mathbf{h}_i \right\|_2}, \\
\mathbf{t}_s &= [\mathbf{h}_i \| \mathbf{p}_{y_i}], \qquad
\mathbf{t}_q = [\mathbf{h}_j \| \mathbf{0}].
\end{aligned}
\end{equation}
This prototypical formulation solves a core challenge. Alternatives like one-hot encoding would result in a \textit{variable token dimension}, while decomposing the task into binary problems would prevent the model from reasoning about \textit{inter-class relationships}. Our approach ensures a consistent token size while enabling the Transformer to reason over all class concepts in a shared context.
% The design of this token structure solves a core challenge for a Transformer-based GFM: it must represent an arbitrary N-way classification task while maintaining a fixed dimensionality for the Transformer. A direct approach, such as pairing an item's embedding with a one-hot label vector, is not viable as the token's dimension would \textit{vary} with the number of classes. Another alternative that achieves a fixed dimension is to decompose the task into N separate binary classifications. However, this is also suboptimal as it prevents the model from reasoning about the crucial relationships \textit{between} classes. Our prototypical formulation resolves both of these issues, using fixed-dimensional prototypes to ensure a consistent token size while enabling the Transformer to reason over all class concepts in a shared context.

\subsection{In-context Reasoning}

With the task syntactically unified into tokens, the In-Context Prediction phase is designed to perform the more complex challenge of \textbf{semantic unification}. This is accomplished by two key components: The ICL Transformer performs the contextual reasoning, learning a task-specific mapping from the prompted examples. This is followed by the Prototypical Head, which provides a dynamic classification mechanism adapting to any N-way task.

\textbf{Transformer-based Reasoning.} The ICL Transformer consists of a stack of $L$ identical layers designed to process the unified set of support and query tokens and produce context-aware embeddings. The design is inspired by the principle of causal attention masking, which has been proven essential for in-context learning on structured tabular data by TabPFN \citep{tabpfn}. This principle ensures the query items do not influence the representation of the support set, nor should they influence each other. We implement this via a specialized \textbf{two-stage process}. 

\paragraph{Stage 1: Context Refinement:} The first stage of each layer builds a rich, task-specific context from the support set. To achieve this, a multi-head self-attention mechanism is applied exclusively to the set of support tokens, $T_{\mathcal{S}}$. This allows the support examples to interact and form a coherent representation of the task's semantics. The output is a set of refined support embeddings, $T'_{\mathcal{S}}$:
\begin{equation}
% \vspace{-0.5em}
T'_{\mathcal{S}} = \text{SelfAttention}(T_{\mathcal{S}})
% \vspace{-0.5em}
\end{equation}
\paragraph{Stage 2: Information Gathering:} This stage uses the refined context to inform the representation of the query tokens. This is the core in-context learning step where the model applies its understanding to the prediction targets. We use a multi-head cross-attention mechanism where the query tokens $T_{\mathcal{Q}}$ serve as the query, while the refined support embeddings $T'_{\mathcal{S}}$ serve as both the keys and values:
\begin{equation}
T'_{\mathcal{Q}} = \text{CrossAttention}(Q = T_{\mathcal{Q}}, K = T'_{\mathcal{S}}, V = T'_{\mathcal{S}})
\end{equation}
% Mark the notation more clearly
As is standard, residual connections and LayerNorm are applied around each sub-module to ensure stable training.

\textbf{Prototypical Prediction.} The final stage of our framework is the Prototypical Head, which performs the final, tuning-free classification. Our architecture is designed to maintain a separation of roles within each token embedding: the ``item space'' serves as the primary input for reasoning, while the ``class space'' serves as the dedicated output space. The ICL Transformer learns to project its final, context-aware prediction into this class-space portion of its output embeddings. Therefore, for the final prediction, \textbf{we use only the class-space portion of the embeddings from the Transformer}. The final classification then proceeds as follows: First, a prototype vector $p_c$ for each class $c$ is computed by taking the element-wise mean of the class-space portions of all final support embeddings. Each query's final class-space embedding is then classified based on its cosine similarity to each class prototype, and these scores are converted into a probability distribution via a softmax function. This mechanism is non-parametric and allows GILT to adapt to any N-way classification task on the fly.

\subsection{Inference-Time Output Refinement}

This stage adds inference-time refinements on top of the shared GILT backbone. We use them to improve prediction-time robustness and, for link prediction, inject structural cues that are known to matter. Our main refinement is test-time augmentation (TTA), motivated by the strong empirical benefits of ensemble in tabular foundation models such as TabPFN \citep{tabpfn}. Specifically, we apply TTA across node, link, and graph classification by constructing views via random feature rotations and averaging their predictions, which improves robustness to noise. For link prediction, prior work has shown that standard MPNNs are limited by the expressive power of 1-WL and therefore do not adequately capture the pairwise information needed to predict a link\citep{seal}. We therefore introduce an MPLP-inspired \citep{mplp} node labeling estimation strategy to strengthen pairwise encoding for link-level tasks, while leaving the shared backbone unchanged.

\subsection{Pre-training}

GILT is not trained to solve specific tasks but is taught the general meta-skill of in-context learning with \textbf{only one unified model}. The goal of pre-training is to optimize its parameters to become an effective few-shot reasoner. \textbf{Compared with many previous GFMs, our pre-training corpus is substantially more diverse} \citep{ofa,graphany,gcope,riemanngfm}. Specifically, it consists of 22 datasets spanning domains such as citation, social, and molecular networks, totaling over 450,000 nodes and 4 million edges, with individual graph sizes ranging from tens to over 170,000 nodes. The test datasets are completely disjoint from the pre-training datasets. GILT learns via a multi-task objective covering node, link, and graph classification, on features with dimensions varying from single digits to over 8,000. Further details are in Appendix \ref{app:pre-training}.

% Add more statistics for pretraining dataset

% The learnable parameters of GILT are optimized through a large-scale, multi-task pre-training designed to teach the model the general skill of in-context learning. We pre-train GILT on over 10 unique public graphs from various domains, including citation networks, social networks, and molecular structures. The training is guided by a multi-task objective that covers node prediction, link prediction, and graph classification.
At each training step, a few-shot task consisting of a support set $\mathcal{S}$ and a query set $\mathcal{Q}$ is generated from our diverse training corpus. This task is then formatted and passed through the GILT architecture to produce predictions. A standard cross-entropy loss is computed between these predictions and the ground-truth labels of the query items, $y_j$:
\begin{equation}
\mathcal{L} = -\frac{1}{|\mathcal{Q}|} \sum_{x_j \in \mathcal{Q}} \log(P(y=y_j | x_j)) 
\end{equation}
This loss is then backpropagated to update all learnable parameters in the model. By optimizing the model over millions of such task instances, the framework is not trained to memorize specific graphs or labels. Instead, it is explicitly trained to learn the meta-skill of \textbf{inferring a task's rules from a given support set and applying them to a query set}, thereby acquiring its ability to perform in-context generalization on completely unseen graphs.

\section{Experiment}

We conduct comprehensive experiments to evaluate GILT across three fundamental graph learning tasks: node classification, link prediction, and graph classification. The core objective is to assess its few-shot performance on unseen graphs against a suite of contemporary GFMs.

\subsection{Experiment Setup}

\textbf{Datasets.} To ensure a fair comparison, we selected datasets that are canonical for their respective tasks in the literature. For \textbf{node classification}, we use the widely-cited \texttt{Cora}, \texttt{Citeseer}, \texttt{Pubmed} \citep{planetoid}, and \texttt{WikiCS} \citep{wikics} benchmarks. For \textbf{link prediction}, we again use the Planetoid datasets and add the \texttt{ogbl-collab} \citep{ogb} benchmark for large-scale evaluation. For \textbf{graph classification}, we employ standard OGB benchmarks \citep{ogb}, \texttt{ogbg-molhiv} and \texttt{ogbg-molpcba}, which are popular benchmarks for this task. Crucially, the features in these standard benchmarks are \textbf{not natural language}, but high-dimensional numerical features. The statistics are shown in Appendix \ref{app:dataset}. 

\textbf{Baselines} We compare GILT against a comprehensive set of baselines representing key paradigms. The competitive landscape differs significantly for each of our three target tasks, so we outline our comparison strategy individually:

% \textbf{Baselines for Node Classification:} For \textbf{node classification}, a rich set of models is available, allowing for a detailed comparison. We group these baselines into two categories to ensure clarity: direct competitors evaluated under the same few-shot protocol, and contextual baselines from other paradigms that provide a broader reference. The first group includes our \textbf{direct few-shot competitors}, which are all evaluated under the same N-way, K-shot protocol.  We further divide this group by their adaptation mechanism. The first sub-group is \textbf{Tuning-Based Models}, which adapt to new tasks by performing gradient-based updates. This includes GCOPE \citep{gcope} and RiemannGFM \citep{riemanngfm}. The second sub-group is \textbf{ICL Models}, which adapt without gradients. This includes OFA \citep{ofa}, which uses a unified prompt graph, and GraphAny \citep{graphany}, which uses an analytical inference method. The second group provides broader context by comparing GILT against \textbf{LLM-based zero-shot baselines}. This includes models like ZeroG \citep{zerog} and GOFA \citep{gofa}, which leverage rich textual descriptions of the target classes. Their mechanism contrasts with our few-shot setting, where class semantics must be inferred exclusively from the provided K-shot examples.

For \textbf{Node Classification}, prior benchmarks are relatively fragmented and often rely on inconsistent evaluation protocols. To ensure a fair comparison, \textbf{we therefore re-evaluate all applicable baselines under a unified few-shot setting}. Our comparisons include standard supervised and self-supervised baselines (MLP, GCN \citep{gcn}, GAT \citep{gat}, DGI \citep{dgi}, and GraphCL \citep{graphcl}) as strong general-purpose anchors, few-shot GFM baselines that adapt through either tuning (GCOPE \citep{gcope}, RiemannGFM \citep{riemanngfm}, and MDGFM \citep{mdgfm}) or in-context inference (OFA \citep{ofa} and GraphAny \citep{graphany}), and LLM-based methods such as LLaga \citep{llaga}, ZeroG \citep{zerog}, and GOFA \citep{gofa}. We treat these LLM-based methods as broader contextual references rather than strictly matched few-shot competitors, since they rely on textual node and class descriptions unavailable in our setting. For \textbf{Link Prediction}, few-shot graph foundation model baselines are extremely scarce: UniLP \citep{unilp} is the only link-level GFM baseline we were able to identify for clean reproduction. Standard GNN link predictors under a strict 5-shot protocol collapse to near-random guessing, so they do not provide a meaningful few-shot reference. We therefore compare GILT primarily against fully supervised link prediction methods, including standard GNNs (GCN \citep{vgae}, GraphSAGE \citep{sage}) and specialized methods (SEAL \citep{seal}, MaskGAE \citep{maskgae}) trained with full training supervision. This fully supervised setting is substantially easier than ours, since those models are optimized in-task with many labeled edges. For \textbf{Graph Classification}, we compare against available few-shot GFM baselines, including OFA \citep{ofa} and GFT \citep{gft}, as well as standard supervised GNNs such as GCN \citep{gcn} and GAT \citep{gat}. Importantly, \textbf{GILT is the only method evaluated in a unified \emph{cross-task} regime (node, link, and graph) with one shared framework, while all baselines are evaluated \emph{in-task} with task-specific setups.} This makes our setting \textbf{strictly harder.} Details for baseline evaluation are shown in Appendix \ref{app:baseline}.

\textbf{Evaluation Protocol} We adhere to a strict evaluation protocol to prevent data leakage. The support set is sampled from the training split, and the query set for final evaluation is sampled from the test split. We report \texttt{Accuracy} for node classification, \texttt{Hits@K} for link prediction and \texttt{ROC-AUC} for graph classification. On the \texttt{Planetoid} datasets, following prior work, we create a random 70\%/10\%/20\% train/valid/test split and report \texttt{Hits@100}; the negative pool for Hits@100 is sampled uniformly and is matched in size to the test positive split. All other evaluations use the official public data splits where available. 

\subsection{Performance and Efficiency Analysis}

\begin{table*}[t]
\centering
\caption{\textbf{Few-shot Node Classification Performance.} We report 1-shot and 5-shot accuracy (\%) against state-of-the-art few-shot baselines. GILT is evaluated in a unified cross-task setting.}
\label{tab:node_classification}
\vspace{-0.5em}

\resizebox{\textwidth}{!}{%
\begin{tabular}{@{}lcccccccccc@{}}
\toprule
& \multicolumn{2}{c}{\textbf{Cora}} & \multicolumn{2}{c}{\textbf{Citeseer}} & \multicolumn{2}{c}{\textbf{Pubmed}} & \multicolumn{2}{c}{\textbf{WikiCS}} & \multicolumn{2}{c}{\textbf{Average}} \\
\cmidrule(lr){2-3} \cmidrule(lr){4-5} \cmidrule(lr){6-7} \cmidrule(lr){8-9} \cmidrule(lr){10-11}
\textbf{Model} & \textbf{1-shot} & \textbf{5-shot} & \textbf{1-shot} & \textbf{5-shot} & \textbf{1-shot} & \textbf{5-shot} & \textbf{1-shot} & \textbf{5-shot} & \textbf{1-shot} & \textbf{5-shot} \\ \midrule
\multicolumn{11}{@{}l}{\textit{Supervised Baselines}} \\
MLP & $31.14 \pm 4.53$ & $42.54 \pm 5.89$ & $26.08 \pm 5.17$ & $45.30 \pm 3.68$ & $48.86 \pm 4.74$ & $62.12 \pm 1.68$ & $28.30 \pm 4.84$ & $55.07 \pm 4.45$ & $33.59$ & $51.26$ \\
GCN & $47.40 \pm 13.37$ & $68.90 \pm 5.85 $ & $29.56 \pm 5.24$ & $58.10 \pm 8.82$ & $57.02 \pm 4.67$ & \underline{$69.88 \pm 3.13$} & $37.53 \pm 5.98$ & $65.85 \pm 3.32$ & 43.00 & 65.56\\
GAT & \underline{$55.12 \pm 11.68$} & $69.90 \pm 3.14$ & $37.76 \pm 8.99$ & $59.08 \pm 7.55$ & \underline{$60.88 \pm 3.80$} & $68.50 \pm 4.73$ & $36.11 \pm 8.42$ & \textbf{67.35 $\pm$ 2.30} & 47.47 & \underline{66.21} \\ 
\midrule
\multicolumn{11}{@{}l}{\textit{Self-Supervised Pre-training}} \\
DGI & $42.70 \pm 7.74$ & $61.44 \pm 4.34$ & $32.82 \pm 7.85$ & $50.78 \pm 6.03 $ & $58.78 \pm 3.89$ & $69.16 \pm 6.51$ & $38.94 \pm 2.64$ & $63.04 \pm 2.78$ & 43.31 & 61.11 \\
GraphCL & $37.18 \pm 8.46$ & $63.32 \pm 4.13$ & $33.08 \pm 6.13$ & $52.24 \pm 3.56$ & $57.58 \pm 4.18$ & $68.20 \pm 4.90$ & $36.38 \pm 8.79$ & $63.68 \pm 2.37$ & 41.06 & 61.86 \\
\midrule
\multicolumn{11}{@{}l}{\textit{Tuning-Based GFMs}} \\
RiemannGFM & $25.08 \pm 9.52$ & $46.82 \pm 15.73 $ & $27.22 \pm 9.21$ & $36.52 \pm 13.56$ & $44.36 \pm 8.48$ & $57.28 \pm 6.68$ & \underline{$49.85 \pm 3.30$} & $53.21 \pm 4.20$ & 36.63 & 48.46 \\
GCOPE & $39.05 \pm 1.73 $ & $67.06 \pm 1.41$ & \textbf{55.87 $\pm$ 0.23} & \underline{$63.90 \pm 1.37$} & $40.89 \pm 1.22$ & $64.34 \pm 1.94$ & $38.72 \pm 0.40 $ & $47.73 \pm 1.29$ & 43.63 & 60.76 \\
MDGFM & $38.80 \pm 8.99$ & $60.94 \pm 2.96$ & $30.26 \pm 8.65$ & $57.88 \pm 9.50$ & $50.64 \pm 10.78$ & $65.90 \pm 2.87$ & OOM & OOM & - & - \\
\midrule
\multicolumn{11}{@{}l}{\textit{ICL Models}} \\
OFA & $30.52 \pm 0.62$ & $ 41.30 \pm 1.89 $ & $40.86 \pm 0.17$ & $52.01 \pm 1.12 $ & $31.07 \pm 0.85$ & $37.70 \pm 0.66$ & $38.50 \pm 1.12$ & $49.23 \pm 0.50$ & 35.24 & 45.06 \\
GraphAny & $49.30 \pm 5.95$ & \underline{72.68 $\pm$ 2.47} & $42.66 \pm 8.30$ & $62.08 \pm 4.98$ & $56.48 \pm 8.98$ & $69.54 \pm 2.75$ & \textbf{51.12 $\pm$ 7.79}  & $57.86 \pm 9.53$ & \underline{49.89} & 65.54 \\
\midrule
\textbf{GILT} & \textbf{56.36 $\pm$ 7.67} &  \textbf{73.22 $\pm$ 3.80}  & \underline{$ 48.35 \pm 4.18 $} & \textbf{66.17 $\pm$ 1.78} &  \textbf{61.74 $\pm$ 6.03}  & \textbf{71.86 $\pm$ 1.55} & $46.13 \pm 1.82$ & \underline{$66.80 \pm 2.38$} & \textbf{53.15} & \textbf{69.51} \\
\bottomrule
\end{tabular}%
}
\vspace{-0.5em}
\end{table*}

In this section, we present the main results, focusing on GILT's two main strengths: its state-of-the-art few-shot performance across tasks and its superior inference efficiency.

% \textbf{Node Classification.} Our main results, presented in Table \ref{tab:main_node_classification}, demonstrate GILT's superior performance in few-shot node classification. Crucially, GILT achieves the highest average accuracy in both the 1-shot (50.12\%) and 5-shot (66.60\%) settings. In the most challenging 1-shot setting on WikiCS, GILT achieves 61.44\% accuracy, significantly outperforming all competitors and highlighting its ability to generalize from minimal context. Furthermore, compared to tuning-based models, GILT's 5-shot performance is superior to RiemannGFM across all datasets and surpasses GCOPE on three of four benchmarks. Crucially, GILT achieves this without any test-time gradient updates, making it far more efficient. While GraphAny shows competitive performance as a specialized analytical solver for node classification, GILT achieves this same competitive standing as a framework designed for generalization across node, link, and graph level tasks. This validates the power of our approach, showcasing that a generalist model can perform on par with a state-of-the-art specialist. 

\textbf{Node Classification.} As shown in Table \ref{tab:node_classification}, GILT sets a new state of the art in few-shot node classification, achieving \textbf{the highest average accuracy} in both 1-shot (53.15\%) and 5-shot (69.51\%) settings, with leading results on most dataset-shot combinations. Importantly, this gain is obtained under \textbf{a significantly stronger setting: GILT is text-free, requires no test-time tuning, and operates as a unified model across node, link, and graph tasks.} Even against strong task-specific baselines, GILT still delivers superior overall performance, highlighting the effectiveness of its unified in-context learning framework. Another instructive pattern is that \textbf{classical supervised and self-supervised GNN baselines outperform several tuning-based GFMs} on many settings under our unified evaluation. We attribute this to the high sensitivity of tuning-based adaptation in the extreme few-shot regime: methods that rely on per-task gradient updates are particularly vulnerable to changes in protocol and target datasets, and therefore do not always transfer cleanly from their original evaluation settings to our unified benchmark. At the same time, the strong performance of these re-evaluated classical baselines suggests that earlier results for standard GNNs were partly constrained by suboptimal hyperparameter choices for few-shot tasks; once a more reasonable combination of hyperparameter is chosen, they become much stronger anchors than is often assumed. In contrast, GILT avoids this fragility altogether, as its in-context adaptation requires no downstream tuning, making it substantially more robust and practical.

\begin{table*}[t]
\vspace{-0.5em}
\caption{Few-shot performance on link prediction and graph classification. GILT is evaluated in a unified cross-task setting.}
\vspace{-0.5em}
\label{tab:link_graph_combined}
\centering

%--- LEFT TABLE (Standard Link Prediction) ---
\begin{minipage}{0.58\linewidth}
    \centering
    \subcaption{Link prediction performance (Hits@K \%). Fully supervised baselines are trained on the train split, while others are evaluated in the 5-shot setting.}
    \label{tab:sub_link_pred}
    \resizebox{\linewidth}{!}{%
    \begin{tabular}{@{}lcccc@{}}
    \toprule
    \textbf{Model} & \textbf{Cora} & \textbf{Citeseer} & \textbf{Pubmed} & \textbf{ogbl-collab} \\
    \midrule
    GCN (Supervised) & $66.79 \pm 1.65$ & $67.08 \pm 2.94$ & $53.02 \pm 1.39$ & $44.75 \pm 1.07$ \\
    SAGE (Supervised) & $55.02 \pm 4.03$ & $57.01 \pm 3.74$ & $39.66 \pm 0.72$ & $48.10\pm 0.81$ \\
    SEAL (Supervised) & $81.71 \pm 1.30$ & $83.89 \pm 2.15$ & \underline{$75.54 \pm 1.32$} & $64.74 \pm 0.43$ \\
    MaskGAE (Supervised) & \underline{$82.48 \pm 0.76$} & \underline{$86.11 \pm 2.26$} & \textbf{80.38 $\pm$ 0.82} & \underline{$65.84 \pm 0.47$} \\
    \midrule
    UniLP (5-shot) & $52.91 \pm 4.62$ & $50.40 \pm 7.49$ & $54.44 \pm 1.34$ & $58.00 \pm 3.10$\\
    \midrule
    \textbf{GILT\tnote{*} (5-shot)} & \textbf{85.29 $\pm$ 2.46} & \textbf{87.68 $\pm$ 0.91} & $72.30 \pm 2.00$ & \textbf{67.83 $\pm$ 0.34} \\
    \bottomrule
    \end{tabular}%
    }
\end{minipage}%
\hspace{0.02\linewidth}% This adds a small gap between the tables
%--- RIGHT TABLE (Cross-Task Link Prediction) ---
\begin{minipage}{0.4\linewidth}
    \centering
    \subcaption{Graph classification performance in AUC under a 5-shot setting.}
    \label{tab:graph_class}
    \resizebox{\linewidth}{!}{%
        \begin{tabular}{@{}lcc@{}}
        \toprule
        \textbf{Model} & \textbf{ogbg-molhiv} & \textbf{ogbg-molpcba} \\
        \midrule
        GCN (5-shot) & $56.23 \pm 6.98$ & $58.43 \pm 1.47$ \\
        GAT (5-shot) & \underline{$56.36 \pm 7.22$} & $51.60 \pm 0.86$ \\
        \midrule
        OFA (5-shot) & $ 49.05 \pm 1.37 $ & $ 51.01 \pm 0.57 $ \\
        GFT (5-shot) & $ 52.78 \pm 7.51 $ &  \textbf{59.67 $\pm$ 1.41}  \\
        \midrule
        \textbf{GILT (5-shot)} & \textbf{58.17 $\pm$ 8.87} & \underline{58.83 $\pm$ 0.79} \\
        \bottomrule
        \end{tabular}
    }
\end{minipage}
\vspace{-1.5em}

\end{table*}

\textbf{Link Prediction.} As shown in Table \ref{tab:link_graph_combined}a, GILT delivers strong few-shot link prediction performance. Compared with fully supervised baselines, which operate in a substantially easier setting than ours,\textbf{GILT still outperforms these baselines on Cora, Citeseer, and ogbl-collab, and achieves the best results on these three benchmarks}. Pubmed remains the main exception, where fully supervised specialist training retains an advantage. These gains indicate that GILT has learned highly generalizable patterns of link formation, which can be rapidly activated by in-context learning from only a few support examples.

\textbf{Graph Classification.} The framework's strong performance extends to the difficult task of few-shot graph classification on challenging molecular benchmarks (Table \ref{tab:graph_class}). Under this 5-shot setting, GILT achieves a ROC-AUC of 58.17\% on \texttt{ogbg-molhiv} and 58.83\% on \texttt{ogbg-molpcba}. It is the strongest method on \texttt{ogbg-molhiv}, and on \texttt{ogbg-molpcba} it remains highly competitive, outperforming GCN, GAT, and OFA while trailing only GFT \citep{gft} which needs dataset specific tuning. These results show that GILT transfers effectively to graph-level tasks even on challenging molecular benchmarks. In particular, its strong performance on the especially large and diverse \texttt{ogbg-molpcba} dataset is achieved with a single unified, tuning-free, cross-task backbone rather than a graph-classification-specific training pipeline. Overall, these results highlight both the difficulty of graph-level transfer and the versatility of GILT as a multi-task foundational model.

\begin{figure}[t]
% \vspace{-1em}
    \centering
    \includegraphics[width=0.8\linewidth]{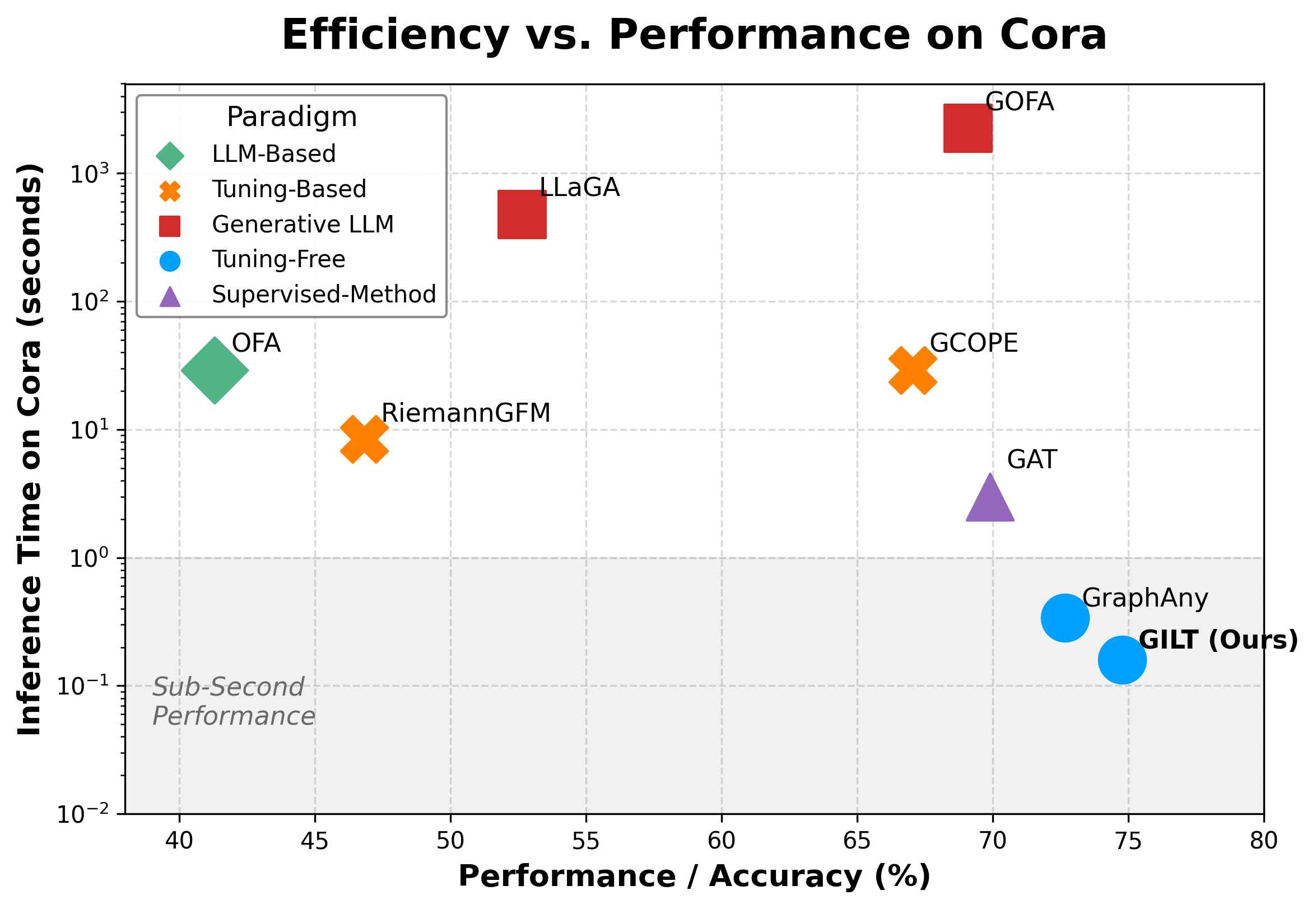}
    \vspace{-1em}
    \caption{Efficiency vs. Accuracy on Cora node classification. The y-axis is the measured time and the x-axis is accuracy. All models are 5-shot, except for the LLM-based zero-shot baselines.}
    \label{fig:speed}
    \vspace{-2em}
\end{figure}

\begin{table*}[t]
% \vspace{-0.5em}
\caption{Ablation study results showing 5-shot accuracy (\%).}
% \vspace{-1em}
\label{tab:ablation_study_full}
\centering
\vspace{-0.5em}
% This table is wider, so resizebox might be needed depending on your margins.
\resizebox{0.82\textwidth}{!}{% 
\begin{tabular}{@{}lcccc@{}}
\toprule
\textbf{Model Variant} & \textbf{Cora} & \textbf{Citeseer} & \textbf{Pubmed} & \textbf{WikiCS} \\ \midrule
\textbf{GILT (Full Model)} & \textbf{73.22 $\pm$ 3.80} & $66.17 \pm 1.78$ & \textbf{71.86 $\pm$ 1.55} & \textbf{66.80 $\pm$ 2.38} \\ \midrule
\multicolumn{5}{l}{\textit{Ablations on the ICL Transformer}} \\
\quad w/o ICL Transformer & $13.00 \pm 0.00$ & $7.70 \pm 0.00$ & $18.00 \pm 0.00$ & $2.51 \pm 0.00$ \\
\quad w/ Full Token for Prediction & $72.97 \pm 3.38 $ & $66.50 \pm 2.02 $ & $70.19 \pm 2.23 $ & $57.19 \pm 3.03$ \\ 
\midrule 
\multicolumn{5}{l}{\textit{Ablations on the Graph Encoder}} \\
\quad w/o Graph Encoder & $57.50 \pm 7.07$ & $52.31 \pm 1.78$ & $60.41\pm 1.10 $ & $36.11 \pm 1.89$ \\
\quad w/ Non-linear GCN & $70.76 \pm 0.93$ & $61.03 \pm 2.50$ & $69.06 \pm 3.03 $ & $50.98 \pm 1.78$ \\
\quad w/ 2-layer Encoder & $70.52 \pm 2.59 $ & \textbf{67.40 $\pm$ 0.08}  & $65.96 \pm 1.97 $ & $55.28 \pm 0.93$ 
\\ \bottomrule
\end{tabular}%
} % End of resizebox
\vspace{-1em}
\end{table*}

\begin{table}[h]
\caption{Effect of inference-time refinement.}
\label{tab:task_bias_ablation}
\centering
\scriptsize
\setlength{\tabcolsep}{4pt}
\vspace{-0.5em}
\begin{tabular}{@{}lcccc@{}}
\toprule
\multicolumn{5}{c}{\textbf{Node classification}} \\
\midrule
\textbf{Variant} & \textbf{Cora} & \textbf{Citeseer} & \textbf{Pubmed} & \textbf{WikiCS} \\
\midrule
base & $68.68 \pm 4.66$ & $61.90 \pm 4.72$ & $71.19 \pm 1.73$ & $60.65 \pm 0.64$ \\
+ TTA & $73.22 \pm 3.80$ & $66.17 \pm 1.78$ & $71.86 \pm 1.55$ & $66.80 \pm 2.38$ \\
\midrule
\multicolumn{5}{c}{\textbf{Link prediction}} \\
\midrule
\textbf{Variant} & \textbf{Cora} & \textbf{Citeseer} & \textbf{Pubmed} & \textbf{ogbl-collab} \\
\midrule
base & $85.78 \pm 0.82$ & $88.95 \pm 0.92$ & $66.52 \pm 5.75$ & $56.08 \pm 0.65$ \\
+ node labeling & $85.29 \pm 2.46$ & $87.68 \pm 0.91$ & $72.30 \pm 2.00$ & $67.83 \pm 0.34$ \\
\bottomrule
\end{tabular}
\vspace{-2.5em}
\end{table}

\textbf{Efficiency Analysis.}
A critical advantage of GILT's tuning-free paradigm is its efficiency, visualized in Figure \ref{fig:speed}. For non-tuning-based methods, the reported time is the total inference time on the Cora test split. For tuning-based methods, we include the dataset-specific tuning time on Cora, since it is part of their deployment. All timings are measured on the same machine using a single NVIDIA RTX 4090 GPU. The plot reveals a stark performance gap across different paradigms. Tuning-free models, such as GILT and GraphAny, are clustered within the sub-second region, demonstrating near-instantaneous response. In contrast, traditional supervised methods like GCN and GAT, while established, still require several seconds for inference. Quantitatively, \textbf{GILT achieves a speedup of approximately 20$\times$ over GAT}. The efficiency gap widens dramatically when compared to adaptation-heavy tuning-based methods and inference-heavy LLM-based methods. \textbf{GILT achieves a speedup of over 180$\times$ compared to the tuning-based method GCOPE, and an unprecedented 14000$\times$ speedup over the generative LLM approach GOFA.} This underscores the practical scalability of the tuning-free ICL approach and positions GILT as a robust solution for real-world, latency-sensitive applications.

\subsection{Architectural Validation and Analysis}
To understand the sources of GILT's strong performance, we conduct a series of analyses to validate its key components and deconstruct its in-context learning behavior.

\textbf{What Architectural Design Makes GILT an Effective In-Context Learner?}
To identify which components are essential for GILT’s in-context learning ability, we perform a series of ablation studies, with results in Table \ref{tab:ablation_study_full}.

The ablation results highlight two main findings. First, the \textbf{ICL Transformer} is indispensable: removing it entirely (\textit{w/o ICL Transformer}) causes a catastrophic performance collapse across all datasets, while using the full token for prediction leads to a smaller but consistent degradation. Second, the \textbf{graph encoder} is also crucial. Removing it causes substantial performance drops, and replacing the deep linear encoder with either a non-linear GCN or a shallow 2-layer encoder generally reduces performance. Overall, these results validate the importance of combining a specialized ICL Transformer with a deep linear graph encoder.

\begin{table}[t]
    \centering
    % \vspace{-1em}
    \caption{Comparison of GILT's 5-shot performance against LLM-based zero-shot baselines.}
    \label{tab:context_zeroshot}
    \vspace{-0.5em} % Optional: Adjusts vertical spacing
    \resizebox{0.5\textwidth}{!}{%
        \begin{threeparttable}
            \begin{tabular}{@{}lcccc@{}}
                \toprule
                \textbf{Model (Setting)} & \textbf{Cora} & \textbf{Citeseer} & \textbf{Pubmed} & \textbf{WikiCS} \\ \midrule
                LLaGA & 52.65 $\pm$ 0.57 & 35.80 $\pm$ 0.02 & 15.84 $\pm$ 0.17 & $52.96 \pm 0.30$ \\
                ZeroG & $58.44 \pm 5.04$ & $49.32 \pm 4.85$ & \underline{$68.13 \pm 7.86$} & $44.79 \pm 10.93$ \\
                GOFA & \underline{69.10 $\pm$ 0.08} & \underline{$62.48 \pm 0.02 $} & 65.91 $\pm$ 0.02 & \underline{66.25 $\pm$ 0.41} \\ 
                \midrule
                \textbf{GILT (5-shot)} & \textbf{73.22 $\pm$ 3.80} & \textbf{66.17 $\pm$ 1.78} & \textbf{71.86 $\pm$ 1.55} & \textbf{66.80 $\pm$ 2.38} \\ \bottomrule
            \end{tabular}%
        \end{threeparttable}
    }
    \vspace{-2em}
\end{table}

We further isolate the effect of these lightweight inference-time refinements and report them separately in Table \ref{tab:task_bias_ablation}. Using node classification as a representative example, we find that TTA consistently improves performance, supporting its role as a robust inference-time enhancement in our pipeline. For link prediction, the additional node-labeling component yields large gains on Pubmed and ogbl-collab but small trade-offs on Cora and Citeseer.

% \begin{figure}
%     \centering
%     \includegraphics[width=\linewidth]{images/few-shot.png}
%     \vspace{-2em}
%     \caption{The influence of the number of shots (K) on GILT's few-shot performance.}
%     \label{fig:few-shot}
%     \vspace{-2em}
% \end{figure}

\textbf{How well does GILT learn from in-context examples?}
To probe the effectiveness of the knowledge GILT acquires from context, we compare it against zero-shot LLMs. These baselines utilize explicit textual descriptions of the classes that often requires laborious pre-processing to obtain, while GILT must infer semantics from the examples alone.

Table \ref{tab:context_zeroshot} show that with an only 5-shot context, GILT’s performance clearly surpasses strong LLM baselines with explicit semantic class labels. This highlights a crucial distinction in their underlying mechanisms: while LLMs perform knowledge retrieval by applying vast, pre-existing linguistic information, GILT performs fundamental, graph-native reasoning. It successfully infers a class's functional definition purely from its numerical and structural context.

\section{Conclusion}

% We presented GILT, a framework designed to be a more universal Graph Foundational Model by operating without textual data or per-graph tuning. The key innovation is a graph-native tokenization pipeline that allows a pre-trained Transformer to perform in-context learning, leveraging task semantics directly from prompted examples at inference time. Our empirical results confirm that this method attains competitive few-shot performance across various unseen benchmarks, outperforming baselines in both generality and inference speed. 

We introduced GILT, a novel Graph Foundational Model designed to be both \textbf{LLM-free} and \textbf{tuning-free}. Our key innovation is reframing few-shot graph tasks as a token-based reasoning problem, allowing a pre-trained Transformer to learn from examples directly at inference time. Experiments confirm that this in-context learning approach achieves strong few-shot performance, offering a more general and efficient solution than prior methods.
\clearpage
\newpage

\bibliography{iclr2026_conference}
\bibliographystyle{icml2026}

\appendix
\section{Dataset Details}
\label{app:dataset}

This section provides detailed statistics for the datasets used in our pre-training corpus and for our downstream evaluations. All evaluation datasets were strictly held out and unseen during the pre-training phase.

\subsection{Pre-training Datasets}

To learn a general-purpose in-context reasoner, GILT was pre-trained on a large and diverse corpus of 22 publicly available graph datasets. This corpus was curated to span multiple domains (social, citation, transportation, web, and molecular) and task levels (node, link, and graph) to ensure the model learns robust and generalizable patterns. A summary of the key datasets included in the pre-training corpus is provided in Table \ref{tab:pretrain_datasets}.

\begin{table*}[t]
\small
\caption{Statistics of datasets used in the GILT pre-training.}
\label{tab:pretrain_datasets}
\centering
\resizebox{\textwidth}{!}{%
\begin{tabular}{@{}llrrrrr@{}}
\toprule
\textbf{Dataset} & \textbf{Domain} & \textbf{\# Graphs} & \textbf{\# Nodes} & \textbf{\# Edges} & \textbf{\# Features} & \textbf{\# Classes/Tasks} \\
\midrule
ogbn-arxiv \citep{ogb} & Citation & 1 & 169,343 & 1,166,243 & 128 & 40 \\
CS \citep{amazon} & Co-authorship & 1 & 18,333 & 163,788 & 6,805 & 15 \\
Physics \citep{amazon} & Co-authorship & 1 & 34,493 & 495,924 & 8,415 & 5 \\
Computers \citep{amazon} & Co-purchase & 1 & 13,752 & 491,722 & 767 & 10 \\
Photo \citep{amazon} & Co-purchase & 1 & 7,650 & 238,162 & 745 & 8 \\
Flickr \citep{graphsaint} & Social & 1 & 89,250 & 899,756 & 500 & 7 \\
USA \citep{struc2vec} & Transportation & 1 & 1,190 & 13,599 & 1,190 & 4 \\
Brazil \citep{struc2vec} & Transportation & 1 & 131 & 1,038 & 131 & 4 \\
Europe \citep{struc2vec} & Transportation & 1 & 399 & 5,995 & 399 & 4 \\
Wiki \citep{pane} & Web pages & 1 & 2,405 & 17,981 & 4,973 & 17 \\
BlogCatalog \citep{pane} & Social & 1 & 5,196 & 343,486 & 8,189 & 6 \\
DBLP \citep{corafull} & Citation & 1 & 17,716 & 105,734 & 1,639 & 4 \\
FacebookPagePage \citep{lastfm} & Social & 1 & 22,470 & 342,004 & 128 & 4 \\
DeezerEurope \citep{lastfm} & Social & 1 & 28,281 & 185,504 & 128 & 2 \\
LastFMAsia \citep{lastfm} & Social & 1 & 7,624 & 55,612 & 128 & 18 \\
bace \citep{moleculenet} & Molecular & 1,513 & 34.1 (avg) & 73.7 (avg) & 9 & 2 \\
bbbp \citep{moleculenet} & Molecular & 2,039 & 23.9 (avg) & 51.6 (avg) & 9 & 2 \\
tox21 \citep{moleculenet} & Molecular & 7,831 & 18.6 (avg) & 38.6 (avg) & 9 & 12 \\
toxcast \citep{moleculenet} & Molecular & 8,575 & 18.8 (avg) & 38.5 (avg) & 9 & 617 \\
clintox \citep{moleculenet} & Molecular & 1,478 & 26.1 (avg) & 55.8 (avg) & 9 & 2 \\
muv \citep{moleculenet} & Molecular & 93,087 & 24.2 (avg) & 51.2 (avg) & 9 & 17 \\
sider \citep{moleculenet} & Molecular & 1,427 & 33.6 (avg) & 70.7 (avg) & 9 & 27 \\ 
\bottomrule
\end{tabular}
} % <-- CLOSE THE RESIZEBOX
\end{table*}

\subsection{Evaluation datasets}

We evaluated GILT's few-shot performance on a suite of 8 unseen benchmark datasets. These datasets were chosen as they are standard in the literature and cover all three primary graph learning tasks. Detailed statistics for each evaluation dataset are summarized in Table \ref{tab:eval_datasets}.

\begin{table*}[t]
\caption{Statistics of datasets used in the GILT evaluation.}
\label{tab:eval_datasets}
\centering
% V-- WRAP THE TABULAR ENVIRONMENT --V
\resizebox{\textwidth}{!}{%
\begin{tabular}{@{}llrrrrr@{}}
\toprule
\textbf{Dataset} & \textbf{Domain} & \textbf{\# Graphs} & \textbf{\# Nodes} & \textbf{\# Edges} & \textbf{\# Features} & \textbf{\# Classes/Tasks} \\
\midrule
Cora \citep{planetoid} & Citation & 1 & 2,708 & 10,556 & 1,433 & 7 \\
Citeseer \citep{planetoid} & Citation & 1 & 3,327 & 9,104 & 3,703 & 6 \\
Pubmed \citep{planetoid} & Citation & 1 & 19,717 & 88,648 & 500 & 3 \\
WikiCS \citep{planetoid} & Web pages & 1 & 11,701 & 216,123 & 300 & 10 \\
ogbl-collab \citep{ogb} & Co-authorship & 1 & 235,868 & 1,285,465 & 128 & Link Prediction \\
ogbg-molhiv \citep{ogb} & Molecular & 41,127 & 25.5 (avg) & 27.5 (avg) & 9 & 1 \\
ogbg-molpcba \citep{ogb} & Molecular & 437,929 & 26.0 (avg) & 28.1 (avg) & 9 & 128 \\ \bottomrule
\end{tabular}%
} % <-- CLOSE THE RESIZEBOX
\end{table*}

\subsection{Introductions to the Datasets}

To make the benchmark coverage explicit, we provide a brief description of each dataset used in this paper.

\paragraph{Pre-training datasets.}
\begin{itemize}[leftmargin=10pt]
    \item \textbf{ogbn-arxiv} \citep{ogb}: a citation network of arXiv papers, where each node is a paper and labels correspond to subject areas.
    \item \textbf{CS} \citep{amazon}: a co-authorship graph in computer science from the Amazon/Coauthor benchmark collection.
    \item \textbf{Physics} \citep{amazon}: a co-authorship graph in physics from the same benchmark suite.
    \item \textbf{Computers} \citep{amazon}: an Amazon co-purchase graph from the computer products category.
    \item \textbf{Photo} \citep{amazon}: an Amazon co-purchase graph from the photo products category.
    \item \textbf{Flickr} \citep{graphsaint}: a social graph with user connectivity and profile-level node features.
    \item \textbf{USA} \citep{struc2vec}: a transportation network (airport-style benchmark) where connectivity reflects routes.
    \item \textbf{Brazil} \citep{struc2vec}: a transportation network from Brazil in the same benchmark family.
    \item \textbf{Europe} \citep{struc2vec}: a transportation network from Europe in the same benchmark family.
    \item \textbf{Wiki} \citep{pane}: a web-page graph benchmark with attributed nodes.
    \item \textbf{BlogCatalog} \citep{pane}: a social network benchmark of bloggers with multi-class labels.
    \item \textbf{DBLP} \citep{corafull}: an academic network benchmark from the citation/co-authorship domain.
    \item \textbf{FacebookPagePage} \citep{lastfm}: a social graph where nodes are Facebook pages and edges capture inter-page links.
    \item \textbf{DeezerEurope} \citep{lastfm}: a user social network from the Deezer music platform.
    \item \textbf{LastFMAsia} \citep{lastfm}: a social graph of LastFM users in Asia with country-level labels.
    \item \textbf{bace} \citep{moleculenet}: a molecular property prediction dataset focused on BACE-1 inhibitor activity.
    \item \textbf{bbbp} \citep{moleculenet}: a molecular dataset for blood-brain barrier penetration prediction.
    \item \textbf{tox21} \citep{moleculenet}: a multi-task molecular toxicity benchmark.
    \item \textbf{toxcast} \citep{moleculenet}: a large-scale multi-task molecular toxicity benchmark with many assay endpoints.
    \item \textbf{clintox} \citep{moleculenet}: a molecular benchmark for clinical trial toxicity and FDA approval-related endpoints.
    \item \textbf{muv} \citep{moleculenet}: a sparse virtual screening benchmark with challenging molecular activity tasks.
    \item \textbf{sider} \citep{moleculenet}: a molecular benchmark for side-effect related prediction tasks.
\end{itemize}

\paragraph{Evaluation datasets.}
\begin{itemize}[leftmargin=10pt]
    \item \textbf{Cora} \citep{planetoid}: a classic citation network benchmark for node classification.
    \item \textbf{Citeseer} \citep{planetoid}: a citation graph benchmark with sparse bag-of-words node features.
    \item \textbf{Pubmed} \citep{planetoid}: a large biomedical citation network benchmark.
    \item \textbf{WikiCS} \citep{wikics}: a Wikipedia-based benchmark for node classification on web pages.
    \item \textbf{ogbl-collab} \citep{ogb}: a collaboration graph benchmark for link prediction.
    \item \textbf{ogbg-molhiv} \citep{ogb}: a molecular graph benchmark for HIV activity prediction.
    \item \textbf{ogbg-molpcba} \citep{ogb}: a large molecular graph benchmark with many bioassay tasks.
\end{itemize}

\section{Baseline Introductions}
\label{app:baseline_intro}

In this section, we briefly introduce all baseline methods used in our experiments. These baselines can be classified into four families: tuning-based methods, in-context learning methods, LLM-based methods, and supervised baselines.

\subsection{Tuning-Based Methods}

\paragraph{GCOPE \citep{gcope}.}
GCOPE operates by projecting disparate graph features into a shared dimension, typically through SVD (singular value decomposition). The mechanism introduces virtual coordinator nodes that form fully connected subnetworks with internal nodes and establish inter-coordinator edges to bridge isolated domains. This architecture creates a joint adjacency matrix intended to unify independent graph distributions into a cohesive system. Training employs a composite objective that combines graph contrastive learning with an auxiliary feature reconstruction loss, aiming to preserve salient data characteristics while distilling transferable representations. Downstream adaptation is achieved by converting target tasks into graph-level problems via induced subgraphs, followed by standard finetuning or the application of graph prompts.

\paragraph{RiemannGFM \citep{riemanngfm}.}
RiemannGFM is a graph foundation model that utilizes a structural vocabulary of trees and cycles to learn universal graph representations. The mechanism is built on a product bundle that integrates Riemannian manifolds for local geometry and tangent spaces for global structural encodings. Its layers comprise a vocabulary learning module that employs cross-geometry attention to update node coordinates in hyperbolic and hyperspherical manifolds. Concurrently, a global learning module utilizes bundle convolution and parallel transport to aggregate node encodings while addressing tangent space incompatibility. Pretraining is conducted via geometric contrastive learning between the hyperbolic and hyperspherical views within a shared tangent space. Downstream tasks are performed by training a classification head on the generated node encodings.

\paragraph{MDGFM \citep{mdgfm}.}
MDGFM addresses cross-domain structural discrepancies by implementing a unified framework centered on topology-aware alignment. The architecture utilizes a decoupled embedding mechanism where an adaptive balance token dynamically weighs node features against aggregated topological information. To mitigate inherent noise and align disparate structures, the model integrates a graph structure learning module that refines source graphs and extracts domain-invariant knowledge. Pre-training is driven by a contrastive learning objective that maximizes mutual information between original and refined graph representations to ensure the preservation of global structural properties. For downstream adaptation, MDGFM employs a dual-stream prompt strategy—combining a meta prompt to model inter-domain relationships with a task-specific prompt for precise target alignment—facilitating robust knowledge transfer even in few-shot scenarios.

\paragraph{GFT \citep{gft}.}
GFT operates by identifying transferable structural patterns as computation trees derived from the graph message-passing process. These computation trees are utilized as tokens within a shared structural vocabulary, allowing the model to encode generalizable patterns across disparate domains and tasks. The mechanism unifies node-, edge-, and graph-level classification by extracting task-relevant subtrees as basic learning instances for pretraining. For downstream tasks, the model undergoes supervised fine-tuning where the entire parameter set is updated using target-specific subtrees to align the learned vocabulary with downstream labels. 

\subsection{In-Context Learning Methods}

\paragraph{OFA \citep{ofa}.}
OFA is a graph foundation model that unifies diverse graph data by representing nodes, edges, and tasks using a standardized natural language description framework. The mechanism utilizes a Graph-to-Text approach where graph structural information and metadata are converted into textual prompts, which are then processed by an LLM backbone. By aligning disparate feature spaces into a common linguistic embedding space, OFA enables a single model to handle multiple graph domains and tasks simultaneously. For downstream tasks, the model employs a prompt-based strategy where task-specific natural language instructions are appended to the input, allowing the model to adapt to new domains through in-context learning.

\paragraph{GraphAny \citep{graphany}.}
GraphAny is a graph foundation model designed for node classification that applies an analytical solution to generalize to any target graph without requiring training on that specific dataset. Its mechanism utilizes a set of graph filters to produce multiple candidate predictions, which are then adaptively combined using weights produced by a lightweight meta-network based on the graph's structural and feature statistics. For downstream tasks, the model identifies the optimal combination of its internal filters and solve an analytical optimization problem using the provided few-shot labels.

\paragraph{UniLP \citep{unilp}.}
UniLP is a universal link prediction foundation model that leverages in-context learning to generalize across diverse graph datasets without task-specific fine-tuning. Its mechanism addresses the challenge of conflicting connectivity patterns by utilizing a set of in-context links sampled from the target graph as reference. These examples, along with the query link, are processed through a shared subgraph GNN encoder and an attention-based fusion module, allowing the model to adaptively recognize the specific structural logic of the current graph. For downstream tasks, UniLP enables zero-shot transfer by simply providing a few in-context examples during inference, achieving performance competitive with or superior to fully supervised, domain-specific models.

\subsection{LLM-Based Methods}

\paragraph{LLaGA \citep{llaga}.}
LLaGA reformulates graph learning by mapping structural information into the input space of LLMs. The framework reorganizes graph nodes and their local neighborhoods into structure-aware sequences, which are then projected into the LLM’s token embedding space via a versatile projector. This approach allows the model to leverage the pre-trained reasoning capabilities of LLMs to interpret complex graph topologies and node attributes simultaneously. LLaGA demonstrates strong generalizability by performing various graph tasks, such as node classification and link prediction, in both supervised and zero-shot settings across diverse domains without requiring task-specific structural modifications.

\paragraph{ZeroG \citep{zerog}.}
ZeroG addresses cross-dataset zero-shot transferability by utilizing LLMs to unify heterogeneous feature spaces and label sets. The framework employs an LLM-based encoder to map diverse node attributes and class semantics into a consistent embedding space, mitigating the challenge of feature misalignment across domains. To capture structural dependencies without requiring task-specific finetuning, ZeroG introduces a prompt-based subgraph sampling module and a relation-aware GNN that generalizes across varied graph topologies. During inference, it performs zero-shot classification by computing the similarity between node-centric subgraph representations and potential class semantic embeddings, enabling effective generalization to unseen datasets and label spaces.

\paragraph{GOFA \citep{gofa}.}
GOFA implements a generative one-for-all framework by interleaving graph-aware layers with a frozen LLM to achieve joint graph-language modeling. The architecture incorporates a symmetry-preserving GNN that extracts structural representations, which are subsequently integrated into the LLM's transformer blocks through a cross-attention mechanism. This design enables the model to process arbitrary graph topologies and textual attributes within a unified generative decoder, supporting a wide range of tasks through natural language prompting. Pre-trained on diverse graph datasets via a next-token prediction objective, GOFA exhibits strong in-context learning and zero-shot generalization capabilities across various graph-structured domains.

\subsection{Supervised Baselines}

\paragraph{GCN \citep{gcn}.}
GCN is a standard message-passing baseline which updates node representations via the propagation rule $H^{(l+1)} = \sigma(\tilde{D}^{-\frac{1}{2}}\tilde{A}\tilde{D}^{-\frac{1}{2}}H^{(l)}W^{(l)})$, utilizing a normalized adjacency matrix $\tilde{A}$ with added self-loops and its degree matrix $\tilde{D}$. For downstream tasks, the model is paired with a task-specific head and undergoes end-to-end supervised training where all parameters are updated on target data.

\paragraph{GraphSAGE \citep{sage}.}
GraphSAGE is a graph neural network that utilizes layer-wise message passing and neighborhood aggregation to generate node embeddings. For downstream tasks, the model typically undergoes supervised end-to-end training where its parameters are iteratively updated to minimize a specific objective function.

\paragraph{GAT \citep{gat}.}
GAT introduces an attention-based neighborhood aggregation mechanism that assigns non-uniform weights to neighbors using coefficients $a_{ij} = \text{softmax}_j(e_{ij})$, where $e_{ij}$ represents the importance of node $j$ to node $i$. This enables the model to focus on the most relevant parts of the local structure without requiring pre-normalized graph statistics. For downstream adaptation, the attention layers and task-specific heads are typically optimized through supervised finetuning on the target domain.

\paragraph{SEAL \citep{seal}}
SEAL is a framework for link prediction that formulates the task as a subgraph classification problem. Its mechanism extracts $h$-hop enclosing subgraphs for each target link and applies a Double Radius Node Labeling (DRNL) scheme to encode the structural roles of nodes relative to the link's endpoints. These labeled subgraphs are then processed by a graph neural network to learn structural signatures that signify link existence. For downstream tasks, the model undergoes supervised training on these extracted subgraphs to map local structural patterns to link presence labels.

\paragraph{MaskGAE \citep{maskgae}}
MaskGAE employs an asymmetric encoder-decoder architecture to perform masked graph modeling. The mechanism involves masking a portion of the input graph edges to reduce redundancy between contrastive subgraph views during training. While a GNN encoder processes only the visible, unmasked subgraph, the decoder utilizes a structure module for edge reconstruction and a degree module for auxiliary node-level degree regression. The framework optimizes a joint objective that combines binary cross-entropy for structural reconstruction with mean squared error for degree prediction. 

\subsection{Self-Supervised Pretraining Baselines}

\paragraph{DGI \citep{dgi}.}
DGI is a self-supervised approach that learns node representations by maximizing the mutual information between local patch representations and a global graph summary. It employs a contrastive objective where the model learns to distinguish between real node-graph pairings and corrupted alternatives generated through a stochastic shuffling mechanism. By forcing the encoder to capture high-level structural properties that are consistent across the entire graph, DGI produces versatile embeddings that can be utilized for various downstream tasks without the need for manual labels.

\paragraph{GraphCL \citep{graphcl}}
GraphCL is a general contrastive learning framework for GNNs that learns unsupervised node or graph representations by maximizing the agreement between different augmented views of the same graph. The framework utilizes four types of graph augmentations—node dropping, edge perturbation, attribute masking, and subgraph extraction—to create diverse correlated views that force the model to learn invariant structural patterns. By employing a contrastive loss in the latent space, GraphCL enables the extraction of robust features that are transferable to various downstream tasks, such as graph classification, without requiring any manual annotations during the pre-training phase.

\section{Implementation Details}

This section provides the specific implementation details for our model, GILT, and all baselines used in the experiments.

\subsection{GILT Model and Pre-training}

\paragraph{Architecture Details.}
The GILT model evaluated in our experiments was configured with the following architecture. The non-parametric encoder projects all node features into a unified dimension of $d=128$. The structural encoder is a 6-layer linear GCN, with each aggregation step followed by a LayerNorm with learnable affine parameters. The ICL Transformer consists of 2 layers, with 4 attention heads per layer and a hidden dimension of $d=1024$ in the feed-forward networks.

\paragraph{More Design Choices.}

Below we summarize several implementation details and fixed design choices used in the final system.

\begin{itemize}[leftmargin=10pt]
    \item Feature Pre-processing: Our feature alignment pipeline is a multi-step process designed to robustly handle graphs with diverse feature dimensions. For graphs with high-dimensional features, we use SVD for dimensionality reduction. For graphs with low-dimensional features, we apply SVD while keeping the same dimensionality and then perform zero padding to match the final model dimension. Finally, the entire processed feature matrix undergoes a crucial column-wise standard scaling, which normalizes each feature dimension independently and prevents padded zeros from distorting the feature statistics.
    \item Graph Encoder: Within the linear GCN encoder, we found two important, and one counter-intuitive, results. First, including the learnable affine parameters in each LayerNorm step was essential for performance; removing them caused a significant loss. Second, we found that adding residual connections between the linear GCN layers, a standard practice in deep networks, was not effective and did not improve results.
    \item Prototype Formulation: For generating the class prototypes, we confirmed that L2-normalizing the vectors after mean pooling is highly beneficial for stabilizing the model. We also experimented with enforcing an additional orthogonality constraint on the prototypes but found it to be less effective than simple normalization.
    \item Test-time augmentation: As described in the main text, we apply test-time augmentation across node, link, and graph classification by constructing five transformed views and averaging the resulting predictions. We use this as a general inference-time robustness mechanism rather than a task-specific module.
    \item Link prediction structural encoding: As described in the main text, we additionally use an MPLP-inspired \citep{mplp} node labeling estimation strategy for link prediction. Conceptually, this can be viewed as a count-based distance encoding of the local common neighborhood around each candidate edge, in the spirit of prior distance-based node labeling schemes such as DRNL and the broader distance-encoding view of structural representation learning \citep{seal,de}. Concretely, it counts how many shared neighbors fall into a small set of canonical relative-distance encodings with respect to the two endpoints, such as $(1,1)$, $(2,1)$, $(1,2)$, $(1,\infty)$, $(\infty,1)$, and $(2,2)$. We estimate these counts using the MPLP estimator, yielding a compact structural summary without explicit enclosing-subgraph extraction.
\end{itemize}

\paragraph{Pre-training.}
\label{app:pre-training}
The pre-training process was conducted for a total of 50 epochs, with each epoch iterating through all tasks sampled from our pre-training corpus. To improve the model's ability to learn from varying amounts of context, we employed a shot decay schedule, where the number of shots, was gradually decayed from an initial 20 down to 5 over the course of training. To enhance model robustness, we applied two forms of data augmentation: feature dropout and edge dropout. The final multi-task loss is a weighted sum of the individual task losses, with a hyperparameter controlling their relative importance. For link prediction tasks specifically, the support set is constructed with a balanced 1:1 ratio between positive and negative edges, while the overall negative sampling ratio used for pre-training remains 3:1.

\subsection{Baseline Setup}
\label{app:baseline}

For our experiments, we made a distinction between re-evaluating baselines and citing established results. Specifically, for the node classification task, we conducted a fresh evaluation of all Tuning-Based and ICL baselines using their official public codebases to ensure a direct and fair comparison under our strict few-shot protocol. For all other results, including baselines on other tasks and the supervised models, performance is reported from their original publications or other well-established literature to ensure consistency with community standards.

\paragraph{Supervised Models.}
  
For MLP, GCN, and GAT, we implemented lightweight supervised baselines under strict few-shot evaluation protocol. All models were trained from scratch on the few-shot samples only, with early stopping based on validation performance, using dropout of 0.5, learning rate of 0.01,  weight decay of $5\times10^{-4}$ and hidden dimension of 128 for MLP and GCN. We report the mean and standard deviation over five random seeds.

\paragraph{Self-Supervised Pretraining Models.}

For DGI and GraphCL, we used a self-supervised pretraining followed by supervised few-shot adaptation protocol. Both methods first pretrained a GCN
backbone on ogbn-arxiv, and then transferred the pretrained backbone to each downstream dataset. During adaptation, the model was fine-tuned on the few-shot samples only, with early stopping based on validation performance. For DGI, we used a standard corruption-based graph infomax objective; for GraphCL, we used feature masking and edge dropout to construct two graph views and optimized an InfoNCE contrastive loss. We used hidden dimension 128, dropout of 0.5, pretraining learning rate of 0.001, fine-tuning learning rate of 0.01, and fine-tuning weight decay of $5\times10^{-4}$. We report the mean and standard deviation over five random seeds.

\paragraph{Tuning-Based Models.}
For GCOPE \citep{gcope}, we used its official implementation and default parameters. To ensure a strict separation between pre-training and evaluation data, we modified its pre-training corpus to exclude the Planetoid datasets. During few-shot adaptation, we followed standard procedure by freezing the GNN backbone and only tuning the prompt module. For RiemannGFM \citep{riemanngfm}, while also using its default parameters, we observed that its original prediction mechanism utilizes external class information. To ensure a fair comparison focused solely on the ability to learn from the provided examples, we replaced its final head with a simple linear classifier that was then tuned on the few-shot support set.

\begin{table*}[t]
\caption{Comparison of the reported pre-training corpora of GILT and prior GFMs based on information disclosed in the original papers.}
\label{tab:pretraining_corpus_template}
\centering
\scriptsize
\setlength{\tabcolsep}{4pt}
\resizebox{\textwidth}{!}{%
\begin{tabular}{@{}p{3.7cm}p{2.8cm}p{3.1cm}p{7cm}@{}}
\toprule
\textbf{Model} & \textbf{Reported Corpus} & \textbf{Task Coverage} & \textbf{Domain Coverage}\\
\midrule
\textbf{GILT} & 22 datasets & Node, link, graph & Citation, social, transportation, web, molecular \\
OFA \citep{ofa} & 3 datasets & Node, link, graph & Citation, knowledge graphs, molecular  \\
GraphAny \citep{graphany} & 1 dataset & Node & Citation \\
GCOPE \citep{gcope} & 9 datasets & Node & Citation, e-commerce, web \\
RiemannGFM \citep{riemanngfm} & 3 datasets & Node, link & Citation, e-commerce, co-authorship \\
\bottomrule
\end{tabular}
}
\end{table*}

\paragraph{ICL Models.} For OFA \citep{ofa} and GraphAny \citep{graphany}, we utilized their official public codebases to evaluate them in our few-shot setting. For OFA, we used the default model parameters and its standard Sentence Transformer for generating text embeddings; the checkpoint was pre-trained on a corpus including ogbn-arxiv. For GraphAny, we also used its default parameters and employed the official model checkpoint pre-trained on the ogbn-arxiv dataset for all node classification evaluations.

\paragraph{LLM-Based Models.} For LLaGA \citep{llaga}, we used the official implementation and followed its hop-token node classification setting. Since the released checkpointsare not trained under the exact zero-shot transfer setting required by our evaluation, we reproduced the projector training ourselves. To avoid target-dataset leakage, we trained the graph-language projector only on source citation graphs and excluded the evaluation dataset from the training corpus. For the graph features, we used SBERT embeddings and generated the required multi-hop propagated embeddings locally, rather than using the released SimTeG feature package, which depends on additional external encoders. For ZeroG \citep{zerog}, we used the official codebase and reproduced the citation-domain zero-shot setting ourselves, since the released repository does not directly provide the exact training protocol used in our evaluation. We trained on source citation graphs only, excluding the target dataset in each run. For GOFA \citep{gofa}, we used the official implementation and the released instruction-tuned checkpoint. For datasets not directly exposed by the GOFA evaluation scripts, such as CiteSeer, we added TAGLAS-compatible dataset wrappers using the data from \cite{llmbp} and kept the original public test split for zero-shot evaluation. For PubMed, full-test evaluation with GOFA required more than 24 hours, so we reported results on a uniformly sampled subset of 10,000 test nodes under the same inference configuration.

\paragraph{Link Prediction Baselines.} Few-shot graph foundation model baselines for link prediction are extremely limited. In practice, UniLP \citep{unilp} is the only dedicated link-level few-shot GFM baseline we were able to identify and evaluate. We do not report 5-shot results for standard supervised GNN link predictors, because under our protocol they degenerate to near-random guessing and therefore fail to provide a meaningful few-shot reference. We therefore compare mainly against fully supervised link prediction baselines trained with the full training split, which is a substantially easier setting than ours because those methods are optimized directly on the target graph with much richer supervision. For these fully supervised baselines, the performance of GCN \citep{gcn}, GraphSAGE \citep{sage} and SEAL \citep{seal} is reported from \cite{buddy}, since it is a widely recognized and authoritative benchmark for link prediction. We therefore use its published results directly rather than introducing additional implementation and tuning differences on our side. We evaluate MaskGAE \citep{maskgae} under its original setting, since its paper reports AUC on the Planetoid datasets. For UniLP, we keep its setup as faithful as possible to the original method, changing only what is necessary to adapt evaluation to our 5-shot protocol and to report Hits@100 on the Planetoid datasets.

\paragraph{Graph Classification Baselines.} For the graph classification baselines, we evaluate GCN \citep{gcn} and GAT \citep{gat} using the same lightweight supervised 5-shot protocol as in node classification: both models are trained from scratch on the provided few-shot graphs only, with the same optimization setup and early stopping strategy. We evaluate OFA \citep{ofa} in the same 5-shot setting while keeping all other configurations unchanged. More broadly, the relatively weak performance of OFA and GFT under our protocol is consistent with the same mismatch discussed for node classification: these methods rely on support-induced prompt or structural templates whose inferred class semantics become unstable under strict few-shot transfer. For GFT \citep{gft}, its original implementation pre-trains on datasets that include downstream datasets and selects fine-tuning samples separately from the few-shot setting. To ensure a fair comparison, we exclude the target test dataset from its pre-training corpus and restrict fine-tuning to the provided few-shot examples. Both methods are trained only for graph classification. 

\subsection{Computational Resources}

All experiments were conducted on a Linux server equipped with 8 NVIDIA RTX 4090 GPUs and dual Intel(R) Xeon(R) Platinum 8370C CPUs @ 2.80GHz (128 logical CPUs in total). Unless otherwise specified, the efficiency measurements in Figure \ref{fig:speed} were obtained on the same machine using a single NVIDIA RTX 4090 GPU. Our implementation is built using PyTorch \citep{pytorch} and PyTorch Geometric (PyG) \citep{pyg}.

\paragraph{Efficiency Measurement Protocol.} To make the efficiency comparison in Figure \ref{fig:speed} as transparent as possible, we measured all methods on the Cora node classification benchmark under the same hardware setting using a single RTX 4090 GPU. For non-tuning-based methods, the reported time is the end-to-end inference time on the Cora test split. For tuning-based methods, we additionally include the required dataset-specific adaptation time on Cora before test inference, since this cost is part of their practical deployment and is typically more substantial than the lightweight training used by simple supervised baselines such as GCN. We do not include dataset I/O time in the measurement, since it depends on the storage environment rather than the method itself. In contrast, dataset preprocessing time is included whenever it is part of the method's standard pipeline. No multi-GPU parallelism was used for this comparison. We therefore interpret the reported time as the total wall-clock cost needed to obtain predictions on the target benchmark under each method's standard usage mode.

\section{Comparison of Pre-training Corpora}
\label{app:pretraining_comparison}

To contextualize the stronger transfer ability of GILT within the graph domain, it is useful to compare its pre-training corpus against those reported for prior GFMs not only in raw scale, but also in task coverage, domain diversity and modality assumptions. Table \ref{tab:pretraining_corpus_template} summarizes these differences in a compact and citation-friendly format, with the comparison restricted to directly relevant graph-domain methods.

Table \ref{tab:pretraining_corpus_template} highlights three patterns that help explain GILT's transfer performance within the graph-domain setting. First, GILT is pre-trained on a substantially larger corpus than the compared prior GFMs, covering 22 datasets, whereas the reported corpora of OFA, GCOPE, and RiemannGFM range from 3 to 9 datasets and GraphAny reports pre-training on a single dataset. Within the current graph-domain literature, this broader scale exposes GILT to a wider variety of feature spaces, graph structures, and task formulations during pre-training.

Second, GILT combines corpus scale with broader task coverage. Among the compared methods, only GILT and OFA explicitly report pre-training across node-, link-, and graph-level tasks, while GraphAny and GCOPE focus only on node-level tasks and RiemannGFM covers node- and link-level tasks. This difference is important in the graph domain because our evaluation targets a unified graph in-context learning setting across multiple task granularities, so broader task diversity during pre-training is likely to improve robustness at test time.

Third, GILT also appears to cover a wider mix of application domains, including citation, social, transportation, web, and molecular graphs. In contrast, the reported pre-training domains of prior GFMs are more concentrated, with several methods focusing primarily on citation-style benchmarks and only limited extension to areas such as e-commerce, knowledge graphs, co-authorship, or molecular data. Taken together, these graph-domain comparisons suggest that GILT's stronger transferability is supported not only by larger pre-training scale, but also by a more diverse combination of tasks and domains relative to prior graph-domain GFMs.

\begin{algorithm}[t]
\caption{Tuning-Free Inference of GILT}
\label{alg:gilt_inference}
\small
\begin{algorithmic}[1]
\Require Graph adjacency $\mathbf{A}$, node features $\mathbf{X}$, support set $\mathcal{S}$, query set $\mathcal{Q}$, target dimension $d$, GCN depth $L_{gcn}$, Transformer depth $L_{tf}$
\Ensure Prediction probability matrix $\mathbf{P}$ for the query set
\If{$\dim(\mathbf{X}) > d$}
    \State $\mathbf{X}' \leftarrow \textsc{SVDProject}(\mathbf{X}, d)$
\Else
    \State $\mathbf{X}' \leftarrow \textsc{ZeroPad}(\textsc{SVDTransform}(\mathbf{X}), d)$
\EndIf
\State $\mathbf{X}' \leftarrow \textsc{ColumnWiseStandardScale}(\mathbf{X}')$
\State $\tilde{\mathbf{A}} \leftarrow \textsc{NormalizeAdjacencyWithSelfLoops}(\mathbf{A})$
\State $\mathbf{H}^{(0)} \leftarrow \mathbf{X}'$
\For{$l = 1$ \textbf{to} $L_{gcn}$}
    \State $\mathbf{H}^{(l)} \leftarrow \textsc{LayerNorm}(\tilde{\mathbf{A}} \mathbf{H}^{(l-1)})$
\EndFor
\State $\mathbf{H} \leftarrow \mathbf{H}^{(L_{gcn})}$
\For{each class $c \in \mathcal{C}$}
    \State $\mathcal{S}_c \leftarrow \textsc{GetSamplesByClass}(\mathcal{S}, c)$
    \State $\mathbf{p}_c \leftarrow \ell_2\text{-}\textsc{Normalize}(\textsc{MeanPool}(\mathbf{H}[\mathcal{S}_c]))$
\EndFor
\State $\mathbf{T}_{\mathcal{S}} \leftarrow \textsc{Concat}(\mathbf{H}[\mathcal{S}], \mathbf{p}_{\mathbf{Y}})$
\State $\mathbf{T}_{\mathcal{Q}} \leftarrow \textsc{Concat}(\mathbf{H}[\mathcal{Q}], \mathbf{0})$
\For{$l = 1$ \textbf{to} $L_{tf}$}
    \State $\mathbf{T}_{\mathcal{S}} \leftarrow \textsc{TransformerBlock}(\mathbf{T}_{\mathcal{S}}, \mathbf{T}_{\mathcal{S}}, \mathbf{T}_{\mathcal{S}})$
    \State $\mathbf{T}_{\mathcal{Q}} \leftarrow \textsc{TransformerBlock}(\mathbf{T}_{\mathcal{Q}}, \mathbf{T}_{\mathcal{S}}, \mathbf{T}_{\mathcal{S}})$
\EndFor
\State $\mathbf{Z}_{\mathcal{S}} \leftarrow \textsc{ExtractClassSpace}(\mathbf{T}_{\mathcal{S}})$
\State $\mathbf{Z}_{\mathcal{Q}} \leftarrow \textsc{ExtractClassSpace}(\mathbf{T}_{\mathcal{Q}})$
\For{each class $c \in \mathcal{C}$}
    \State $\mathbf{v}_c \leftarrow \ell_2\text{-}\textsc{Normalize}(\textsc{MeanPool}(\mathbf{Z}_{\mathcal{S}}[\mathcal{S}_c]))$
\EndFor
\For{each query sample $j \in \mathcal{Q}$}
    \For{each class $c \in \mathcal{C}$}
        \State $\mathbf{Sim}[j, c] \leftarrow \langle \mathbf{Z}_{\mathcal{Q}}[j], \mathbf{v}_c \rangle$
    \EndFor
    \State $\mathbf{P}[j, :] \leftarrow \textsc{Softmax}(\mathbf{Sim}[j, :])$
\EndFor
\State \Return $\mathbf{P}$
\end{algorithmic}
\end{algorithm}

\section{Pseudo-code of Tuning-Free Inference}
\label{app:pseudocode}

Algorithm \ref{alg:gilt_inference} summarizes the inference pipeline of GILT, including feature pre-processing, graph-native structural encoding, task tokenization, in-context reasoning, and non-parametric prediction.

\section{Hyperparameter Settings}

This section details the hyperparameter settings for GILT. To efficiently find a robust configuration, we performed a hyperparameter search using Bayesian optimization. The search was conducted on a held-out set of validation tasks sampled from our pre-training datasets to identify a single, robust set of parameters. The final optimal values, listed in Table \ref{tab:hyperparameters}, were then frozen and used for all reported experiments across all datasets and tasks.

\begin{table*}[t]
\caption{Hyperparameter search space and optimal values used for the GILT model.}
\label{tab:hyperparameters}
\centering
\small
\begin{tabular}{@{}lll@{}}
\toprule
\textbf{Hyperparameter} & \textbf{Search Space} & \textbf{Optimal Value} \\
\midrule
\multicolumn{3}{l}{\textit{Pre-training Hyperparameters}} \\
\quad Learning Rate & {[1e-6, 1e-3]} & 1e-5 \\
\quad Weight Decay & {[1e-6, 1e-2]} & 1e-4 \\
\quad Model Dropout & {[0, 0.5]} & 0.2 \\
\quad Total Training Epochs & {10, 50, 100} & 50 \\
\midrule
\multicolumn{3}{l}{\textit{Multi-Task Training Details}} \\
\quad Batch Size (Node Tasks) & \{2048, 4096, 8192, 16384\} & 2048 \\
\quad Batch Size (Link Tasks) & \{4096, 8192, 16384, 32768\} & 16384 \\
\quad Batch Size (Graph Tasks) & \{1024, 2048, 4096\} & 1024 \\
\midrule
\multicolumn{3}{l}{\textit{GILT Architecture Hyperparameters}} \\
\quad Unified Dimension (d) & {64, 128, 256} & 128 \\
\quad GCN Encoder Layers & {[2, 8]} & 6 \\
\quad Transformer Layers (L) & {[1, 5]} & 2 \\
\bottomrule
\end{tabular}
\end{table*}

\section{Investigation into the Influence of Shot Number}
\label{app:shot_influence}

\begin{figure}[t]
    \centering
    \includegraphics[width=\linewidth]{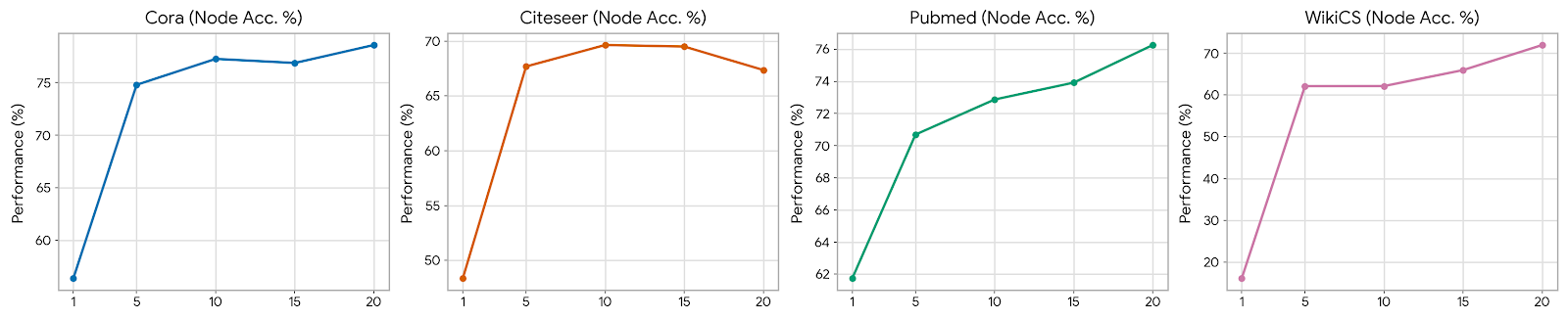}
    \caption{The influence of the number of shots (K) on GILT's few-shot performance. The x-axis represents the number of support examples per class, and the y-axis represents the classification accuracy on the test set. Each line corresponds to a different dataset.}
    \label{fig:few-shot-appendix} % Using a new label for appendix figure
\end{figure}

To better understand how GILT utilizes contextual information, we analyzed its performance across a varying number of support examples (shots) for the node classification task. The results are presented in Figure \ref{fig:few-shot-appendix}.

As illustrated in the figure, there is a clear positive correlation between the number of shots provided in the context and the model's overall performance. The most significant gains in accuracy are typically observed when increasing the shot number from one to ten. As more examples are added to the context, the performance continues to improve, though the rate of improvement gradually diminishes. This analysis confirms that providing a richer context with more examples is an effective method for enhancing GILT's predictive accuracy, which is consistent with the expected behavior of an in-context learning framework.

\section{LLM Usage} 

We utilized LLMs to assist in the preparation of this work. Specifically, we used LLMs for debugging code snippets and for proofreading and improving the clarity of the manuscript's text. The authors reviewed and edited all LLM-generated content and take full responsibility for the final submission.

\end{document}